\documentclass{article}

\PassOptionsToPackage{numbers, compress}{natbib}

  \usepackage[main, final]{neurips_2026}

\usepackage[utf8]{inputenc} %
\usepackage[T1]{fontenc}    %
\usepackage{hyperref}       %
\usepackage{url}            %
\usepackage{booktabs}       %
\usepackage{amsfonts}       %
\usepackage{nicefrac}       %
\usepackage{microtype}      %
\usepackage[table, dvipsnames]{xcolor}         %

\usepackage{graphicx}
\usepackage{subcaption}
\usepackage{tikz}
\usetikzlibrary{positioning}
\usepackage{tabularx}
\usepackage{multirow}
\usepackage{wrapfig}
\usepackage{tabularx}
\usepackage{booktabs}
\usepackage{wrapfig}
\usepackage{graphicx}
\usepackage{colortbl}
\usepackage{caption}

\usepackage{amsmath}
\usepackage{amssymb}
\usepackage{mathtools}
\usepackage{amsthm}

\usepackage[framemethod=TikZ]{mdframed}
\usepackage{tcolorbox}
\definecolor{chart}{HTML}{1f77b4}
\mdfsetup{%
    middlelinecolor =   none,
    middlelinewidth =   1pt,
    backgroundcolor =   blue!5,
    roundcorner     =   5pt,
}
\newtcolorbox{example}[1][]{
  colback=chart!5!white,
  colframe=chart,
  floatplacement=floating,
  title=\centering \textsf{#1}
}

\usepackage{xspace}
\newcommand{\eg}{\textit{e.g.}\xspace}
\newcommand{\ie}{\textit{i.e.}\xspace}
\newcommand{\ours}{\textsc{ReFINE}\xspace}
\newcommand{\oursfull}{Reinforced Fast Weights via Next Sequence Prediction\xspace}

\newcommand{\zhiwei}[1]{}
\newcommand{\sanghyuk}[1]{}
\newcommand{\will}[1]{}

\usepackage{minitoc}
\usepackage{titletoc}

\theoremstyle{plain}

\theoremstyle{definition}

\theoremstyle{remark}

\usepackage[capitalize]{cleveref}
\Crefname{table}{Tab.}{Tabs.}
\crefname{lemma}{lemma}{lemmas}
\Crefname{lemma}{Lemma}{Lemmas}
\crefname{proposition}{proposition}{propositions}
\Crefname{proposition}{Proposition}{Propositions}
\crefname{theorem}{theorem}{theorems}
\Crefname{theorem}{Theorem}{Theorems}
\crefname{corollary}{corollary}{corollaries}
\Crefname{corollary}{Corollary}{Corollaries}
\crefname{assumption}{assumption}{assumptions}
\Crefname{assumption}{Assumption}{Assumptions}
\crefname{remark}{remark}{remarks}
\Crefname{remark}{Remark}{Remarks}

\usepackage[disable,textsize=tiny]{todonotes}

\title{Reinforced Fast Weights via Next-Sequence Prediction 
}

\author{
    Hee Seung Hwang$^{1}\thanks{Equal contribution}$ \quad
    Xindi Wu$^{1*}$
    \quad
    Sanghyuk Chun$^{1}$
    \quad
    Olga Russakovsky$^{1}$ 
    \quad
    Zhiwei Deng$^{2}$ \\ \\
    $^{1}$Princeton University \quad $^{2}$Google DeepMind
}

\begin{document}

\maketitle

\begin{abstract}
    Fast weight architectures offer a promising alternative to standard transformers for long-context modeling by replacing the KV-cache with a recurrently updated fixed-size memory. Despite this architectural shift, they are typically trained with the same next-token prediction (NTP) objective as standard transformers. This creates a mismatch: fast weight models rely on an evolving memory state to support future predictions, while NTP provides only token-level supervision for the immediate next token. We address this mismatch with next-sequence prediction (NSP), a sequence-level extension of NTP that trains models to produce coherent multi-token continuations from their recurrent memory state. To optimize this sequence-level objective, we propose \ours (\textbf{Re}inforced \textbf{F}ast we\textbf{I}ghts via \textbf{N}ext s\textbf{E}quence prediction). \ours selects informative token positions based on prediction entropy, generates multi-token rollouts, assigns sequence-level rewards, and optimizes the model with Group Relative Policy Optimization (GRPO). \ours is applicable throughout the training lifecycle of pre-trained language models: mid-training, post-training, and test-time training. Experiments on LaCT-760M and DeltaNet-1.3B show that \ours consistently outperforms NTP-based supervised fine-tuning in needle-in-a-haystack retrieval, long-context question answering, and diverse tasks in LongBench. \ours provides an effective and versatile framework for improving long-context modeling in fast weight architectures. Code is available at \url{https://github.com/princetonvisualai/ReFINE}.
\end{abstract}

\section{Introduction}
Long-context modeling has become essential for large language models (LLMs). Tasks such as long-document understanding~\cite{shaham2022scrolls, dong2023bamboo}, many-shot in-context learning~\cite{agarwal2024many, li2024long}, and code generation~\cite{liu2023repobench, nichols2024can} require models to extract, store, and reuse information from contexts spanning thousands of tokens~\cite{yen2024helmet,bai2024longbench, bai2025longbench}. While attention-based transformers demonstrate strong performance on these tasks, their computational costs scale quadratically~\cite{keles2023computational} with context length, creating a fundamental bottleneck for both training and inference.

\begin{figure}[th]
    \centering
    \includegraphics[width=\linewidth]{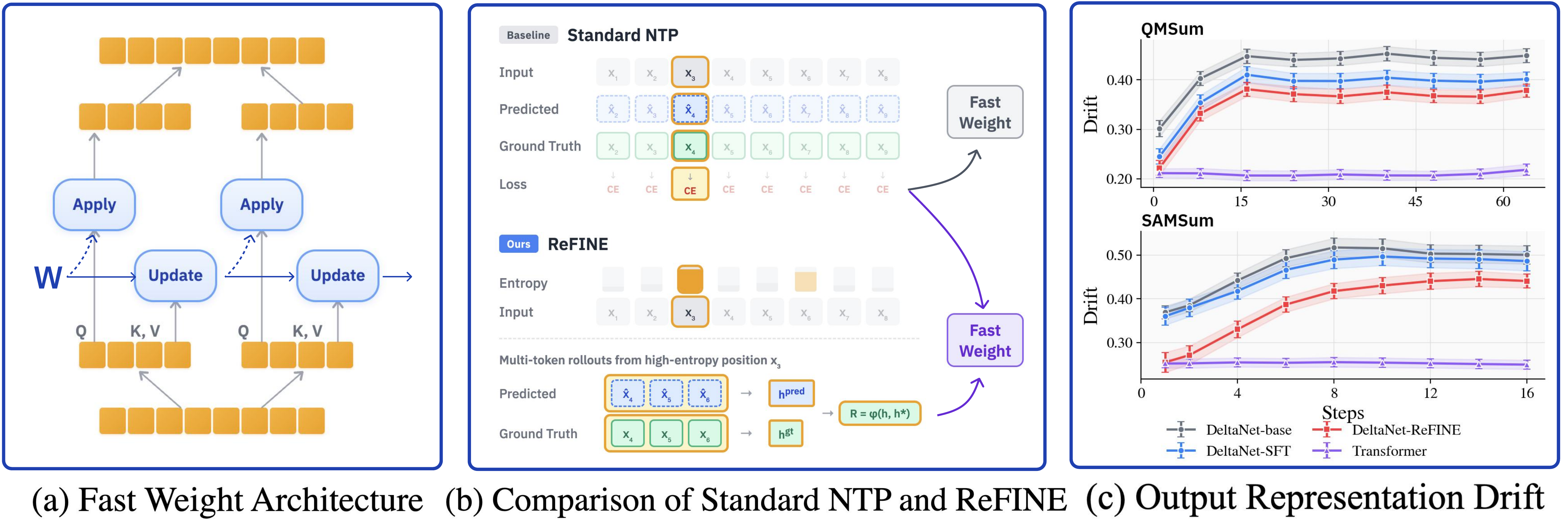}
    \vspace{-15pt}
    \caption{
        \textbf{(a)} Adapted from~\citet{zhang2025test}. Fast weight models replace the KV-cache in standard transformers with a recurrent fixed-size memory ($W$), updated according to~\cref{eq:update}. \textbf{(b)} Standard NTP provides token-level supervision through cross-entropy loss at each position. \ours provides sequence-level supervision by optimizing the rewards of multi-token rollouts at high-entropy positions with RL. \textbf{(c)} Fast weight models show larger divergence in the last layer's output representations, measured by Euclidean distance between predicted and ground truth response trajectories, than standard transformers. \ours substantially reduces this divergence compared to SFT. Error bars show $\pm 1\sigma$ across 50 independent generations. 
    }
    \vspace{-15pt}
    \label{fig:teaser}
\end{figure}

Fast weight architectures offer a promising alternative by addressing the scaling problem through structural changes to the transformer block. Models such as DeltaNet~\cite{yang2024parallelizing}, GatedDeltaNet~\cite{yang2024gated}, and LaCT~\cite{zhang2025test} replace the KV-cache in standard transformers with a dynamically updated, fixed-size memory that learns input-dependent KV mappings on-the-fly (\cref{fig:teaser}a). This design enables efficient inference with constant memory overhead~\cite{sun2024learning}.

However, compressing past tokens into a bounded memory via recurrent updates poses a unique challenge for fast weight models. Unlike transformers, which retain explicit representations of past tokens in the KV-cache, fast weights continually adapt as new tokens are processed. This recurrent compression makes the memory susceptible to forgetting~\cite{schlag2021linear, chaudhry2019tiny} and error accumulation~\cite{ross2011reduction, ranzato2015sequence, schwarz2018progress, li2025mva}, as small deviations in the autoregressive trajectory can compound over time. This requires the training objective of fast weight models to supervise the long-horizon stability of the memory state~\cite{dalal2025one}. 

Despite their architectural differences, fast weight models are pre-trained with the same next-token prediction (NTP) objective used for standard transformer LLMs~\cite{sun2024learning,behrouz2024titans,behrouz2025atlas,yang2024parallelizing,yang2024gated,zhang2025test}. In this work, we argue that NTP is a \textit{suboptimal} objective for fast weight models. NTP provides supervision through teacher-forced ground truth tokens at each position, but does not evaluate the quality of the recurrent memory state after it evolves over multiple steps on the model's own predictions~\cite{gloeckle2024better}. 

To improve the adaptive capacity of fast weights over long horizons, we propose the \textbf{next-sequence prediction (NSP)} objective as a sequence-level extension of NTP. NSP encourages a model to predict a semantically coherent sequence of future tokens conditioned on a given prefix. This objective evaluates whether the information stored in fast weights enables accurate continuation over multiple steps, providing a more appropriate training signal than standard NTP. However, training with NSP introduces two key challenges: (i) standard cross-entropy loss does not naturally extend to multi-token prediction, and (ii) generating sequences for every prefix is computationally prohibitive. We address these challenges by formulating NSP as a reinforcement learning (RL) problem and propose \ours (\oursfull), a \textit{phase-agnostic} training framework for fast weight models. \ours focuses on informative regions of the context and optimizes for NSP using policy gradient updates. 

In a motivating experiment, we compare how autoregressive generation trajectories diverge from ground truth trajectories in fast weight and transformer models. Fast weight models exhibit substantially larger divergence in output representations than transformers, suggesting that recurrent memory updates may be more sensitive to compounding prediction errors. Training with \ours reduces this divergence compared to NTP-based SFT, supporting the role of sequence-level supervision in stabilizing long-horizon generation (\cref{fig:teaser}c).
 
Our experiments also show that \ours improves performance in multiple stages of the language model training lifecycle (\cref{tab:training_phases}): (i) \textit{mid-training}, where \ours reinforces long-context adaptation of fast weights; (ii) \textit{post-training}, where \ours reinforces task-specific adaptation of fast weights; and (iii) \textit{test-time training}, where \ours reinforces prompt-specific adaptation of fast weights. Concretely, applying \ours on LaCT-760M notably improves the average performance on RULER~\cite{hsieh2024ruler} long-context QA tasks by 8.5\% (mid-training), 15.3\% (post-training), and 9.5\% (test-time training), compared to pure NTP training with SFT. For DeltaNet-1.3B, we observe 20.3\% (mid-training), 11.0\% (post-training), and 15.0\% (test-time training) gains with \ours.

\textbf{Our contributions are:}
\begin{itemize}
    \item We introduce the NSP objective for fast weight models to address the lack of sequence-level feedback in standard NTP.
    \item We propose an RL framework to optimize NSP, combining entropy-based token selection and sequence-level rewards.
    \item We demonstrate that \ours improves fast weight models across mid-training, post-training, and test-time training, with 25\% and 8\% average gains over SFT mid-training on LongBench~\cite{bai2024longbench} tasks.
\end{itemize}

\section{Background}

\paragraph{Fast Weight Architectures.}
Fast weight architectures replace the explicit KV-cache in standard transformers with a fixed-size, dynamically updated memory represented by weight matrices. To compress contextual information into the memory, fast weight models recurrently update these weight matrices as new tokens are processed. 
This perspective connects fast weight models to test-time adaptation~\cite{behrouz2024titans} and meta-learning~\cite{clark2022meta}, as the memory is updated continually and task-agnostically during inference. The update rule for fast weights can be generalized as follows~\cite{zhang2025test}:
{
\begin{equation}
W_{t+1} \leftarrow W_t - \eta \nabla_{W_{t}} \ell \bigl(W_{t}k_t, v_t\bigr)
\label{eq:update}
\end{equation}
}where $W$ denotes the fast weight, $\eta$ is the learning rate, and $k_t,v_t$ are the key, value representations of the input token at position $t$. This update rule can be viewed as online learning of a key-to-value mapping. The output representation is retrieved from fast weights via an apply operation, \ie, $o_t = W_t q_t$, where $o_t, q_t$ are the token's output and query representations. \cref{fig:teaser}a illustrates the internal structure of fast weight language models.

\paragraph{Training Phases for Language Models.} 
Pre-trained language models undergo additional training stages that rely on different sources of data and supervision (\cref{tab:training_phases}). 
We follow the standard taxonomy of three additional training phases: \textit{mid-training}, \textit{post-training}, and \textit{test-time training}.

\begin{wraptable}{r}{0.52\columnwidth}
    \vspace{-12pt} %
    \centering
    \small
    \caption{\textbf{Training Phases for Fast Weight Language Models.} \ours can be applied across all phases beyond pre-training to improve long-context modeling, using different data sources.}
    \vspace{-5pt}
    \label{tab:training_phases}
    \begin{tabularx}{\linewidth}{@{}l|X@{}}
      \toprule
      \textbf{Training Phase} & \textbf{Source of Training Data} \\
      \midrule
      Pre-training & General web-scraped data \\
      Mid-training & Similar data to pre-training \\
      Post-training & Task / instruction / preference data \\
      Test-time training & Test data \\
      \bottomrule
    \end{tabularx}
    \vspace{-18pt}
\end{wraptable}

\textit{Mid-training} is an extension (or continued version) of pre-training, generally used for the adaptation of a pre-trained model to specific domains or capabilities \citep{gururangan2020dont}. In this paper, we apply \ours to mid-train models on the same training dataset as pre-training to adapt pre-trained models to the NSP objective.

\textit{Post-training} fine-tunes pre-trained models to follow instructions and align their responses with human preferences~\citep{ouyang2022training, rafailov2024direct, guo2025deepseek}. This phase typically involves SFT on task-specific instruction-response pairs, with relatively fewer gradient steps compared to pre-training. We apply \ours during post-training through the nested learning~\cite{behrouz2025nested} technique: within each training loop, we first use \ours to update the model on the instruction prompt alone, and then use SFT to fine-tune the model's final response.

\textit{Test-Time Training} (TTT) adapts model parameters at inference time using self-supervised objectives to handle distribution shifts from source to target \citep{sun2020test,wang2021tent,gandelsman2022testtime,sun2024learning}. TTT naturally integrates with fast weight architectures by design: each input token updates the fast weights via gradient-based rules, enabling the model to memorize and adapt to the given context on-the-fly~\citep{zhang2025test, behrouz2024titans}. 
However, fast weight models trained purely with NTP can still struggle on long-context retrieval (e.g., Needle-in-a-Haystack~\citep{hsieh2024ruler}) due to unstable fast weight updates or insufficient long-horizon supervision.

\paragraph{RL for Language Modeling.}
Recent work has formulated NTP as a reward maximization problem with RL~\cite{dong2025reinforcement, hatamizadeh2025rlp}. These methods elicit reasoning traces before predicting the next token and assign rewards based on the similarity between the prediction and the ground truth tokens. However, existing approaches focus on standard transformer LLMs that demonstrate basic reasoning behavior. It is still an open question whether RL can be applied to pre-trained fast weight models that lack such reasoning capabilities. We show that RL can improve fast weight models on long-context tasks without prior instruction tuning. See Appendix~\S\ref{app:related} for more discussion of related work.

\begin{figure*}[t!]
    \centering
    \includegraphics[width=\textwidth]{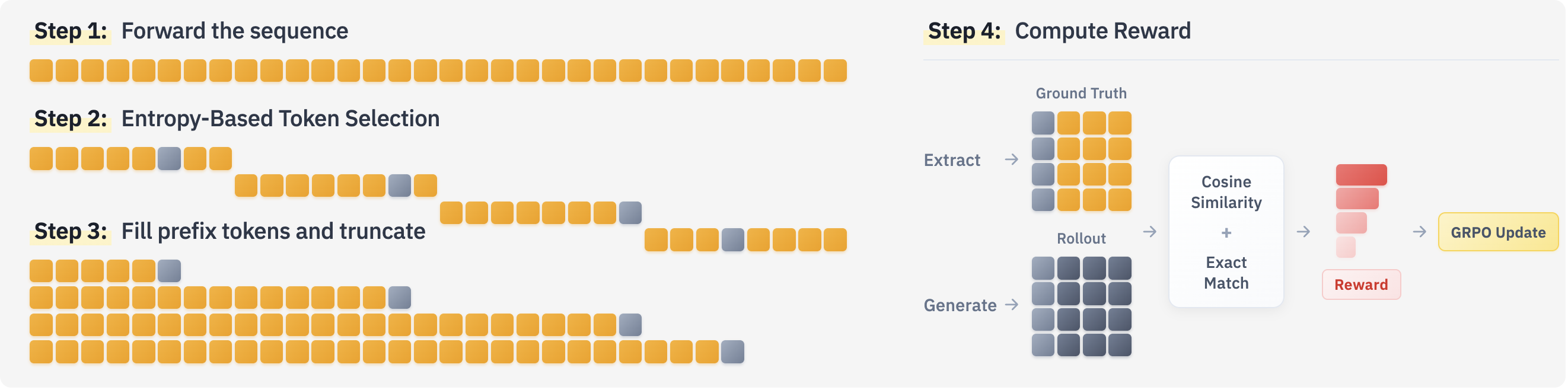}
    \vspace{-10pt}
    \caption{\textbf{\ours.} We forward the sequence through the policy model and compute token-level entropy values, which we use to sample a target token position from each equal-length chunk in the sequence (\textbf{Entropy-Based Token Selection}). Prefixes are copied from the original sequence up to each target token, and the policy model predicts continuations (\textbf{Rollout Generation}). Reward is computed based on the generated rollouts and ground truth tokens (\textbf{Reward Assignment}). Finally, we update the policy model with GRPO (\textbf{Optimization with RL}).
    }
    \label{fig:framework}
\end{figure*}

\section{Method}

Our goal is to improve fast weight models in long-context settings via the NSP objective. We present RL as a solution to the lack of sequence-level feedback in SFT.

\subsection{From Next-Token to Next-Sequence Prediction}
\label{sec:formulation} 

\paragraph{Next-Token Prediction (NTP).} Standard language model pre-training involves minimizing the cross-entropy (CE) loss of the NTP objective. Given an input sequence $S = (x_1, \ldots, x_T)$, the CE loss is computed using the predicted probability of the ground truth token at each position:
{
\begin{equation}
\mathcal{L}_{\mathrm{NTP}}
= \sum_t -\log p(x_{t+1} \mid x_{\le t}).
\end{equation}
}CE loss has two key limitations in long-context settings. First, each term in the summation only considers single-token predictions, ignoring the semantic relationships among the tokens of a multi-token continuation. Second, NTP aggregates the terms uniformly and thus ignores local regions in the sequence that may be especially useful over long contexts~\cite{fang2024wrong}.

\paragraph{Next-Sequence Prediction (NSP).} We propose the NSP training objective for fast weight models to overcome the limitations of standard NTP.
While NTP optimizes token-by-token predictions, NSP optimizes multi-token sequence alignment at selected positions $\mathcal T^* \subseteq \{1, \ldots, T\}$:
{
\begin{equation}
\label{eq:nsp}
\mathcal{L}_{\mathrm{NSP}}
= \sum_{t \in \mathcal{T}^*} \mathcal{L}_{\text{seq}}(\hat{x}_{t+1:t+k}, x_{t+1:t+k}), \, \, \, \,  k > 1
\end{equation}
}where $\mathcal{L}_{\text{seq}}$ measures the discrepancy between predicted ($\hat{x}_{t+1:t+k}$) and ground truth ($x_{t+1:t+k}$) continuations given prefix $x_{\le t}$. 

A straightforward choice for $\mathcal{L}_{\text{seq}}$ is the CE loss. However, naively applying the CE loss at every token position $t$ requires generating $k$-token completions given all possible prefixes, which is computationally impractical for long contexts. Furthermore, directly matching a single reference would over-penalize plausible answers that do not exactly match the ground truth.
For example, ``automobiles move quickly'' and ``cars are fast'' are semantically equivalent, but may incur a high CE loss. 

We tackle this issue with two approaches. First, instead of unrolling tokens at each $t$, we select informative positions $\mathcal{T}^*$ with high NTP entropy, indicating high uncertainty.
Second, we optimize \cref{eq:nsp} with RL to maximize the expected self-supervised reward $R$ of sequence predictions. Let $\pi_{\theta}$ denote the language model parameterized by $\theta$. We define the sequence-level loss as follows:
{
\begin{equation}
\label{eq:seq_rl}
\mathcal{L}_\mathrm{seq}
= -\mathbb{E}_{\hat{x}_{t+1:t+k} \sim \pi_{\theta}(\cdot \mid x_{\leq t} )}\!\big[ R(\hat{x}_{t+1:t+k}, x_{t+1:t+k} ) \big].
\end{equation}
}This formulation has two advantages: (1) optimizing $k$-token predictions leverages higher information content than single-token predictions; (2) we can assign smooth rewards to multiple plausible continuations based on their semantic similarity to the ground truth. For brevity, we use $R(t)$ to denote $R(\hat{x}_{t+1:t+k}, x_{t+1:t+k})$.

\subsection{\ours}
\label{sec:framework}

Our \ours framework (\cref{fig:framework}) consists of four key steps: (1) entropy-based token selection, (2) rollout generation, (3) reward assignment, and (4) optimization with RL.

\paragraph{Entropy-Based Token Selection.} We forward the input sequence $S = (x_1, \ldots, x_T)$ through the policy model $\pi_\theta$ and compute the NTP entropy $H_t = H\!\left(\pi_\theta(\cdot \mid x_{\leq t-1})\right)$ for each token position $t$. We smooth the entropy distribution within $S$ using 1-D average pooling with kernel size $w$. 
We partition $S$ into $c$ contiguous chunks of equal length: $S = (S_1, S_2, \ldots, S_c)$. For each chunk $S_i$, we sample one token position weighted by the softmax of its entropy. Concretely, we compute:
{
\begin{equation}
\label{eq:ent_sampling}
p_i(t) = 
\frac{\mathrm{e}^{\tilde{H}_t}}
{\sum_{t' \in \mathcal{T}_i } \mathrm{e}^{\tilde{H}_{t'}}},
\qquad \mathcal{T}_i = \{ t' \mid x_{t'} \in S_i \}
\end{equation}
}where $\tilde{H}_t$ is the smoothed $H_t$. We then sample an index $t_i \sim p_i(t)$. This produces one entropy-weighted token position for each chunk, yielding a set of sampled positions $\mathcal{T}^* = \{t_1, \ldots, t_c\}$. These positions represent regions that exhibit relatively high local uncertainty, allowing training to focus on challenging predictions spread evenly across the context. 

\paragraph{Rollout Generation.}

For each high-entropy position $t_i \in \mathcal{T}^*$, we construct a truncated prefix $x_{\leq t_i}$, yielding $c$ distinct partial sequences $\{x_{\leq t_1}, \ldots, x_{\leq t_c}\}$ from $S$.
These prefixes capture the full context leading up to each sampled high-entropy position. From each truncated prefix $x_{\leq t_i}$, we generate a $k$-token continuation $\hat{x}_{t_i + 1: t_i + k}$ using the current policy. We extract the hidden states of the predicted and ground truth continuations from the final layer before the logits:
{
\begin{equation}
\mathbf{h}^{\text{pred}}_k(t_i) = \big( \mathbf{h}^{\text{pred}}(t_i+1), \ldots, \mathbf{h}^{\text{pred}}(t_i + k) \big), \qquad \mathbf{h}^{\text{gt}}_k(t_i) = \big( \mathbf{h}^{\text{gt}}(t_i+1), \ldots, \mathbf{h}^{\text{gt}}(t_i + k) \big).
\end{equation}
}We measure the discrepancy between $\mathbf{h}^{\text{pred}}_k(t_i)$ and $\mathbf{h}^{\text{gt}}_k(t_i)$ to compute the reward. 

\paragraph{Reward Assignment.} Given the hidden states of predicted and ground truth continuations $\mathbf{h}_{k}^{\text{pred}}(t_i), \mathbf{h}_k^{\text{gt}}(t_i) \in \mathbb{R}^{k \times d}$, we assign a smooth reward $R$ for an arbitrary similarity function $\varphi$:
{
\begin{equation}
\label{eq:cs_reward}
R^{\varphi}_k(t_i) = \frac{1}{k} \sum_{j=1}^{k} \varphi\big(\mathbf{h}^{\text{pred}}(t_i + j), \mathbf{h}^{\text{gt}}(t_i + j)\big).
\end{equation}
}We use cosine similarity for $\varphi$.
This reward encourages the model to generate hidden state trajectories that align with the ground truth, while remaining fully self-supervised. The purpose of a representation similarity-based reward is to improve contextual generalization and training stability. It assigns smooth, non-zero rewards to semantically similar tokens that lead to hidden state embeddings that are closer in the latent space~\cite{park2024geometry}. Qualitative examples of $R^{\varphi}$ can be found in Appendix~\S\ref{app:qualitative}.

\paragraph{Optimization with RL.} Once we compute the reward for each rollout, we have a set of rollouts $ o_{\mathcal{T}^*} = \{\hat{x}_{t_1+1:t_1+k}, \dots, \hat{x}_{t_c+1:t_c+k} \}$ and corresponding rewards $\mathcal{R}_{\mathcal{T}^*} =  \{R_k^{\varphi}(t_1),\dots,R_k^{\varphi}(t_c)\}$. The rewards from the same sequence $S$ are standardized to compute the advantage following \citet{shao2024deepseekmath}.

We employ the GRPO algorithm \citep{shao2024deepseekmath} to compute the NSP loss based on the rollouts and their relative advantages.
The policy gradients therefore maximize the following objective: 
{
\begin{equation}
\mathcal{J}(\theta) = \mathbb{E}_{ x_{\leq t} \sim \mathcal{D},\, \hat{x}_{t+1:t+k} \sim \pi_{\theta}(\cdot\,|\,x_{\leq t})}[R^{\varphi}_k(t)],
\end{equation}
}where $\mathcal{D}$ is the set of all $\{x_{\leq t}\}_{t=1}^{T}$. To prevent catastrophic forgetting, the final loss is a weighted sum of the NSP loss and the standard NTP loss (computed over the entire sequence $S$), with coefficients $\lambda_{\text{RL}}$ and $\lambda_{\text{SFT}}$ respectively. The weights are adjusted based on the training phase.

\subsection{Hybrid Reward for RL}
\label{sec:hybrid_reward}

TTT introduces unique constraints. First, evaluations are usually conducted in the low-data regime, which leads to smaller batch sizes and limited room for meta-adaptation across episodes. Second, effective memorization of the given context becomes more important than contextual generalization.
For scenarios that require stronger context memorization (\eg, TTT), we introduce a binary exact match reward $R^\text{binary}$ defined as:
{
\begin{equation}
R^{\text{binary}}_k(t_i) = \frac{1}{k} \sum_{j=1}^{k} \mathcal{I}[x_{t_i+j} = \hat{x}_{t_i+j}],
\end{equation} 
}where $\mathcal{I}[\cdot]$ is the indicator function. During post-training, we use the hybrid reward $R^{\text{hybrid}}_k(t_i) = R^{\varphi}_k(t_i) + R^{\text{binary}}_k(t_i)$, since the training and test datasets have similar distributions. The hybrid reward aims to balance contextual generalization and memorization.

\begin{table*}[t!]
    \vspace{-5pt}
    \centering    
    \caption{
    \textbf{Performance on Long-Context Retrieval Tasks.} 
    We evaluate mid-trained (\textbf{MidTr}) models on the NIAH tasks in RULER at 4K, 8K, and 16K context lengths (standard SFT vs. \ours). Highest scores in each category are highlighted in \textbf{bold}.
    } 
    \vspace{-5pt}
    \resizebox{\textwidth}{!}{
        \begin{tabular}{cc | c c c c | c c c c | c c c c | c c c c}
\toprule
& 
& \multicolumn{4}{c|}{RULER S-NIAH} 
& \multicolumn{4}{c|}{RULER MK-NIAH} 
& \multicolumn{4}{c|}{RULER MQ-NIAH} 
& \multicolumn{4}{c}{RULER MV-NIAH} \\

& MidTr
& 4K
& 8K
& 16K
& Avg
& 4K
& 8K
& 16K
& Avg 
& 4K
& 8K
& 16K
& Avg
& 4K
& 8K
& 16K
& Avg \\
            
\midrule

\multirow{3}{*}{\shortstack{LaCT\\760M}}
& --
& 98.8	
& 91.2	
& 95.8
& 95.3
& 70.2	
& 44.8	
& 24.2
& 46.2 
& 39.2		
& 24.9	
& 17.4
& 27.1
& 40.6	
& 27.0	
& 17.7
& 28.4 \\

& SFT 
& 98.4	
& 90.8	
& \textbf{97.6}
& 95.6
& \textbf{70.6}	
& 45.0	
& 24.2
& 46.6
& \textbf{40.7}	
& 25.2	
& 17.4
& 27.8
& \textbf{41.7}	
& 26.3	
& 17.7
& 28.5 \\

& \cellcolor{gray!20}\ours
& \cellcolor{gray!20}\textbf{99.0}	
& \cellcolor{gray!20}\textbf{93.0}	
& \cellcolor{gray!20}96.8	
& \cellcolor{gray!20}\textbf{96.3}	
& \cellcolor{gray!20}70.4	
& \cellcolor{gray!20}\textbf{46.0}	
& \cellcolor{gray!20}\textbf{26.6}	
& \cellcolor{gray!20}\textbf{47.7}	
& \cellcolor{gray!20}40.5	
& \cellcolor{gray!20}\textbf{25.8}	
& \cellcolor{gray!20}\textbf{18.0}	
& \cellcolor{gray!20}\textbf{28.1}	
& \cellcolor{gray!20}41.2	
& \cellcolor{gray!20}\textbf{29.5}	
& \cellcolor{gray!20}\textbf{18.6}	
& \cellcolor{gray!20}\textbf{29.8} \\
            
\midrule
\multirow{3}{*}{\shortstack{DeltaNet\\1.3B}}
& --
& \textbf{100.0}	
& \textbf{100.0}	
& \textbf{100.0}
& \textbf{100.0}
& 23.6	
& 17.8	
& 3.4
& 14.9
& 27.7	
& 3.9	
& 1.8
& 11.1
& 23.9	
& 5.4	
& 2.0
& 10.4 \\

& SFT 
& \textbf{100.0}
& \textbf{100.0}
& \textbf{100.0}
& \textbf{100.0}
& 23.8	
& 19.2	
& 7.8	
& 16.9
& 33.2
& 16.8	
& 3.5
& 17.8
& 28.3	
& 19.3	
& 3.7	
& 17.1 \\

& \cellcolor{gray!20}\ours 
& \cellcolor{gray!20}\textbf{100.0}	
& \cellcolor{gray!20}\textbf{100.0}	
& \cellcolor{gray!20}99.8	
& \cellcolor{gray!20}99.9	
& \cellcolor{gray!20}\textbf{25.0}	
& \cellcolor{gray!20}\textbf{21.4}	
& \cellcolor{gray!20}\textbf{8.8}	
& \cellcolor{gray!20}\textbf{18.4}	
& \cellcolor{gray!20}\textbf{37.3}
& \cellcolor{gray!20}\textbf{18.8}	
& \cellcolor{gray!20}\textbf{3.6}	
& \cellcolor{gray!20}\textbf{19.9}
& \cellcolor{gray!20}\textbf{29.0}	
& \cellcolor{gray!20}\textbf{20.4}	
& \cellcolor{gray!20}\textbf{4.0}	
& \cellcolor{gray!20}\textbf{17.8} \\

\bottomrule
\end{tabular}
}
\label{tab:ruler_niah}
\end{table*}

\begin{table*}[t]
    \centering
    \caption{\textbf{Performance on Multi-Doc QA Tasks.} We evaluate various training strategies during mid-training (\textbf{MidTr}), post-training (\textbf{PostTr}), and test-time training (\textbf{TTT}) for multi-document question and answering tasks.
    For Nested SFT and Nested \ours, we train the model with the training method on the prompt portion of the sample as described in \S\ref{experiment_post_training}.
    Higher is better. First and second highest scores on each task are highlighted in \textbf{bold} and \underline{underline}, respectively.
    }
    \vspace{-5pt}
    \resizebox{\textwidth}{!}{
        \begin{tabular}{cccc | c c c c | c c c c }
\toprule
&&&& \multicolumn{4}{c|}{RULER SQuADQA} 
& \multicolumn{4}{c}{RULER HotpotQA} \\
& MidTr & PostTr & TTT
& 4K
& 8K
& 16K
& Avg
& 4K
& 8K
& 16K
& Avg \\ 
\midrule
\multirow{8}{*}{\shortstack{LaCT\\760M}}
& - & - & -
& 17.5	
& 14.0
& 6.5
& 12.7
& 20.5
& 14.5
& 12.0
& 15.7 \\

& SFT & - & -
& 19.5	
& 14.0
& 6.0
& 13.2
& 21.5
& 15.0
& 12.0
& 16.2 \\

\rowcolor{gray!20}
\cellcolor{white}
& \ours & - & -
& 20.5	
& 15.5
& 8.0
& 14.7
& 23.0
& 16.0
& 12.5
& 17.2 \\

& \ours & SFT & -
& 30.5	
& 20.0
& 7.5
& 19.3
& 25.0
& 16.5
& 11.5
& 17.7 \\

& \ours & Nested SFT & -
& 38.0	
& 20.0
& 7.5
& 21.8
& 24.0
& 16.0
& 12.5
& 17.5 \\

\rowcolor{gray!20}
\cellcolor{white}
& \ours & Nested \ours & -
& \underline{43.5}	
& \underline{24.5}
& \underline{8.5}
& \underline{25.5}
& \underline{27.0}
& 19.5
& \underline{13.0}
& 19.8 \\

& \ours & Nested \ours & SFT
& \underline{43.5}
& 21.5
& \underline{8.5}
& 24.5
& 26.5
& \underline{24.0}
& \underline{13.0}
& \underline{21.2} \\

\rowcolor{gray!20}
\cellcolor{white}
& \ours & Nested \ours & \ours
& \textbf{45.5}	
& \textbf{25.5}
& \textbf{10.0}
& \textbf{27.0}
& \textbf{28.5}
& \textbf{25.5}
& \textbf{15.0}
& \textbf{23.0} \\

\midrule
\multirow{8}{*}{\shortstack{DeltaNet\\1.3B}}
& - & - & -
& 11.0	
& 5.0
& 2.0
& 6.0
& 10.0
& 3.0
& 2.5
& 5.2		 \\

& SFT & - & -
& 9.0	
& 6.0
& 3.0
& 6.0
& 9.0
& 8.0
& 4.5
& 7.2	  \\

\rowcolor{gray!20}
\cellcolor{white}
& \ours & - & -
& 10.0	
& 6.5
& 4.0
& 6.8
& 12.0
& 9.5
& 5.5
& 9.0	  \\
    
& \ours & SFT & -
& 11.0	
& 7.0
& 5.0
& 7.7
& 14.0
& \underline{11.0}
& 6.5
& 10.5	\\

& \ours & Nested SFT & -
& 12.0	
& 8.0
& 5.0
& 8.3
& \underline{16.5}
& 10.0
& 7.5
& 11.3 \\

\rowcolor{gray!20}
\cellcolor{white}
& \ours & Nested \ours & -
& 14.0	
& \underline{10.5}
& \underline{6.5}
& 10.3
& 15.0
& \underline{11.0}
& \textbf{8.5}
& \underline{11.5} \\

& \ours & Nested \ours & SFT
& \underline{16.5}
& \underline{10.5}
& \underline{6.5}
& \underline{11.2}
& 16.0
& 10.5
& 6.5
& 11.0 \\

\rowcolor{gray!20}
\cellcolor{white}
& \ours & Nested \ours & \ours
& \textbf{17.5}
& \textbf{12.5}
& \textbf{7.0}
& \textbf{12.3}
& \textbf{18.0}
& \textbf{13.5}
& \underline{8.0}
& \textbf{13.2} \\

\bottomrule
\end{tabular}
}
\label{tab:ruler_qa}
\end{table*}

\begin{table*}[t]
    \centering
    \small
    \caption{\textbf{Performance on Long-Context Tasks in LongBench.} We study the impact of the learning algorithm during mid-training (\textbf{MidTr}) and test-time training (\textbf{TTT}) on tasks with long-context. \textbf{SFT} denotes the supervised fine-tuning with next-token prediction.
    We evaluate on 12 tasks in LongBench, filtered for samples with at most 16K tokens.
    Details are similar to \cref{tab:ruler_qa}.
    }
    \vspace{-5pt}
    \resizebox{\textwidth}{!}{
        \begin{tabular}{c c c | c c c | c c | c c | c c c | c c | c}
            \toprule
             &&
             & \multicolumn{3}{c|}{Single-doc QA} 
             & \multicolumn{2}{c|}{Multi-doc QA} 
             & \multicolumn{2}{c|}{Summarization}
             & \multicolumn{3}{c|}{Few-shot QA}
             & \multicolumn{2}{c|}{Coding}
             & \\
& MidTr & TTT 
& NQ
& QR
& MF
& HP
& 2W
& QM
& MN
& SS
& TC
& TQ
& LC
& RP
& Avg \\
\midrule
\multirow{5}{*}{\shortstack{LaCT\\760M}}
& - & -
& \underline{6.5}
& 10.5
& 7.2
& 11.7
& 9.8
& 13.6	
& 9.2	
& 14.2
& 10.5
& 8.0
& 26.7
& 29.7
& 13.1 \\

& SFT & - 
& 5.8
& 10.1
& 7.4
& 12.6
& 9.2
& 13.1	
& 10.5	
& 13.8
& 7.5
& 12.2
& 29.8
& 30.1
& 13.5 \\

& \cellcolor{gray!20}\ours & \cellcolor{gray!20}-
& \cellcolor{gray!20}\underline{6.5}
& \cellcolor{gray!20}11.1	
& \cellcolor{gray!20}12.6
& \cellcolor{gray!20}\textbf{19.6}	
& \cellcolor{gray!20}18.0	
& \cellcolor{gray!20}\textbf{14.3}
& \cellcolor{gray!20}\underline{15.9}
& \cellcolor{gray!20}\underline{17.0}
& \cellcolor{gray!20}11.0
& \cellcolor{gray!20}\underline{12.9}
& \cellcolor{gray!20}\textbf{32.9}
& \cellcolor{gray!20}31.1
& \cellcolor{gray!20}16.9 \\

&\ours & SFT
& 5.1	
& \underline{12.6}
& \underline{13.5}
& 15.9
& \underline{18.4}
& 13.2
& \textbf{16.2}
& \textbf{17.4}
& \underline{12.5}
& \textbf{16.2}
& 32.0
& \underline{31.4}
& \underline{17.0} \\

\rowcolor{gray!20}
\cellcolor{white}
& \ours & \ours
& \textbf{6.7}
& \textbf{14.5}	
& \textbf{14.1}	
& \underline{18.4}
& \textbf{22.8}	
& \underline{13.9}
& \underline{15.9}
& \textbf{17.4}	
& \textbf{15.5}	
& 11.8	
& \underline{32.2}
& \textbf{32.3}	
& \textbf{18.0} \\

\midrule
\multirow{5}{*}{\shortstack{DeltaNet\\1.3B}}
& - & -
& \underline{6.5}
& 8.7	
& 10.0	
& 4.8	
& 6.4	
& 12.4	
& \textbf{16.3}	
& 9.4	
& 17.7	
& 15.1	
& 33.8	
& 29.0 
& 14.2 \\

& SFT & -
& 5.7	
& 9.4	
& 8.3	
& 9.2
& \underline{8.6}	
& 14.7	
& 16.1	
& 15.5	
& 22.9	
& 15.2	
& 33.1	
& 29.2 
& 15.7 \\

& \cellcolor{gray!20}\ours & \cellcolor{gray!20}-
& \cellcolor{gray!20}\underline{6.5}
& \cellcolor{gray!20}\underline{9.5}
& \cellcolor{gray!20}10.1	
& \cellcolor{gray!20}\textbf{9.6}
& \cellcolor{gray!20}\textbf{8.7}
& \cellcolor{gray!20}\textbf{15.2}	
& \cellcolor{gray!20}15.0	
& \cellcolor{gray!20}16.2	
& \cellcolor{gray!20}25.0	
& \cellcolor{gray!20}\underline{21.4}	
& \cellcolor{gray!20}\textbf{35.9}	
& \cellcolor{gray!20}\textbf{31.1}	
& \cellcolor{gray!20}\underline{17.0} \\

& \ours & SFT
& 7.2
& \textbf{9.6}
& \underline{10.5}
& 6.8
& 7.2
& \underline{14.9}
& \underline{16.2}
& \textbf{17.3}
& \underline{28.0}
& 16.6
& 34.1
& 29.2
& 16.5 \\

\rowcolor{gray!20}
\cellcolor{white}
& \ours & \ours
& \textbf{7.5}	
& 9.2	
& \textbf{11.5}	
& \underline{9.5}
& \underline{8.6}
& 14.7	
& 16.1	
& \underline{16.5}
& \textbf{31.5}	
& \textbf{24.7}	
& \underline{35.2}
& \underline{30.0}
& \textbf{17.9} \\

\bottomrule
\end{tabular}
}
\label{tab:longbench}
\end{table*}

\section{Experiments}
\label{sec:experiments}

We conduct experiments to show that \ours improves fast weight models on long-context tasks. We first establish the models, datasets, and baselines used for training and evaluation (\S\ref{experiment_setup}). We then report results across three training phases: mid-training (\S\ref{experiment_mid_training}), post-training (\S\ref{experiment_post_training}), and test-time training (\S\ref{experiment_test_time_training}). Finally, we provide analysis on fast weight drift, reward assignment, and entropy-based token selection (\S\ref{sec:analysis}), followed by ablations on rollout length and the number of chunks (\S\ref{sec:ablations}). 

\subsection{Setup}
\label{experiment_setup}

\paragraph{Models.}
We use two pre-trained fast weight language models, LaCT-760M~\cite{zhang2025test} and DeltaNet-1.3B~\cite{yang2024parallelizing}. LaCT adapts the model by updating its fast weight parameters, whereas DeltaNet keeps parameters fixed but updates a parallelizable memory state.
We show that \ours can improve these distinct fast weight mechanisms during mid-training, post-training, and test-time training.

\paragraph{Baselines.} We compare \ours with SFT in our main evaluations. We provide the full \ours training configurations for each phase in Appendix~\S\ref{app:configuration}. We also evaluate additional baselines that train with SFT-based multi-token objectives: (1) up-weighting the CE loss of multiple tokens at high entropy positions~\cite{fang2024wrong}, and (2) enabling multi-token prediction via architectural modification~\cite{gloeckle2024better}. We show that \ours outperforms these baselines in Appendix~\S\ref{app:additional_analysis}, \cref{tab:ruler_niah_baseline,tab:ruler_qa_baseline,tab:longbench_baseline}.

\paragraph{Datasets and Benchmarks.}
For mid-training, we employ a dataset similar to that used to pre-train the fast weight models. Specifically, \textit{we perform mid-training on Long-Data-Collections}~\cite{longdatacollections}, which is the pre-training dataset for LaCT \citep{zhang2025test}.
We evaluate the quality of mid-trained models on RULER NIAH tasks~\cite{hsieh2024ruler} and Booksum~\cite{kryscinski2022booksum}. We consider two additional scenarios: (1) multi-doc QA and (2) broader long-context tasks. For multi-doc QA, we \textit{post-train on synthetically generated SQuADQA and HotpotQA tasks from RULER}. Then, we apply TTT during evaluation. For broader long-context tasks, we employ 12 tasks from LongBench~\cite{bai2024longbench} and apply TTT on the mid-trained models. More details on these datasets can be found in Appendix~\S\ref{app:datasets},~\cref{tab:datasets}.

\subsection{Impact of \ours on Mid-Training}
\label{experiment_mid_training}

We mid-train the models on 200M tokens from Long-Data-Collections~\cite{longdatacollections} for 1 epoch. The models are evaluated on four NIAH tasks in RULER~\cite{hsieh2024ruler}
(Single-key-query-value, Multi-key, Multi-query, and Multi-value)
at 4K, 8K, and 16K context lengths using Language Model Evaluation Harness~\cite{eval-harness}.

\cref{tab:ruler_niah} shows that the \ours mid-trained models consistently outperform the original pre-trained model and the SFT mid-trained model on different tasks. For example, \ours significantly improves DeltaNet in the Multi-key NIAH (+23.5\% from no mid-training and +8.8\% from SFT mid-training). This suggests that \ours leads to improvements in long-context retrieval.

\begin{figure*}[t!]
    \centering
    \vspace{-10pt}
    \begin{subfigure}[b]{0.48\textwidth}
        \centering
        \caption{DeltaNet-1.3B
        }
        \vspace{-3pt}
         \includegraphics[width=\linewidth]{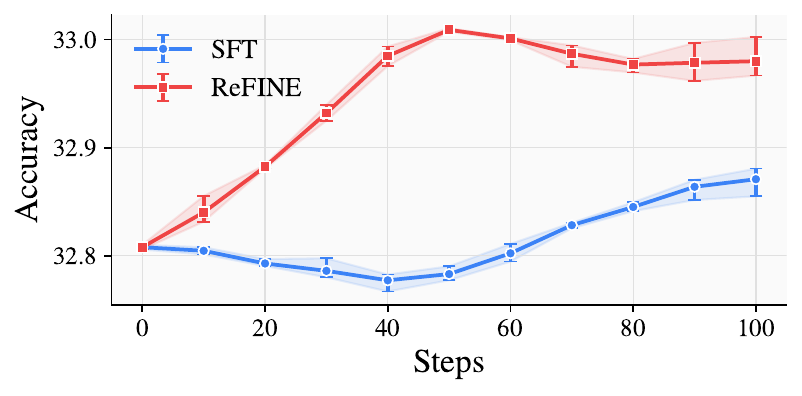}
        \label{fig:ntp_acc_deltanet}
    \end{subfigure}
    \hfill
    \begin{subfigure}[b]{0.48\textwidth}
        \centering
        \caption{LaCT-760M}
        \vspace{-3pt}
        \includegraphics[width=\linewidth]{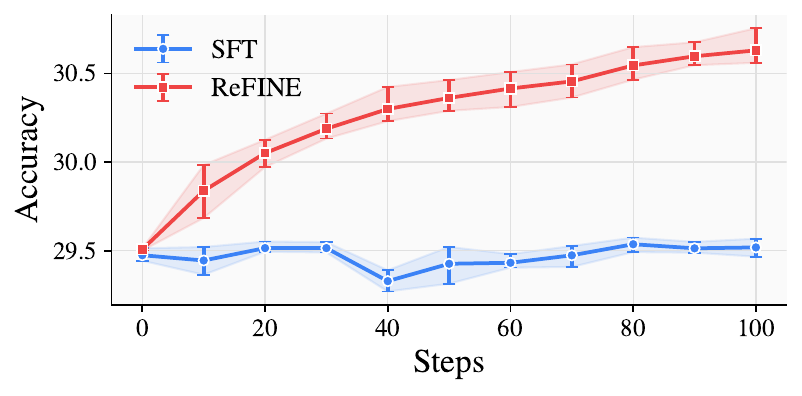}
        \label{fig:ntp_acc_lact}
    \end{subfigure}
    \vspace{-15pt}
    \caption{
        \textbf{NTP Accuracy on Booksum.} \ours mid-training on DeltaNet-1.3B \textbf{(a)} and LaCT-760M \textbf{(b)} consistently improves NTP accuracy on the validation dataset while SFT is stagnant. The error bars show the minimum and maximum values from three independent trials. 
    }
    \vspace{-10pt}
    \label{fig:ntp_acc}
\end{figure*}

We also report the validation NTP accuracy on the Booksum~\cite{kryscinski2022booksum} dataset. \cref{fig:ntp_acc} shows the evolution of validation NTP accuracy during mid-training.
Interestingly, \ours improves NTP accuracy more than SFT even though the SFT baseline directly optimizes NTP. This holds for LaCT, whose NTP accuracy stagnates for SFT, suggesting that NSP provides learning signals that NTP does not.

We report the effectiveness of \ours mid-training on multi-doc QA and LongBench~\cite{bai2024longbench} tasks in \cref{tab:ruler_qa} and \cref{tab:longbench}. In \cref{tab:ruler_qa}, \ours mid-trained models (3rd rows) outperform SFT mid-trained models (2nd rows) by large margins. For example, \ours mid-training improves the average performance of pre-trained DeltaNet on RULER HotpotQA by 25.0\% compared to SFT mid-training. \cref{tab:longbench} shows a similar result; \ours (3rd rows) steadily outperforms SFT (2nd rows).

\subsection{Impact of \ours on Post-Training}
\label{experiment_post_training}

We show that \ours strengthens post-training, which aims to align the model's responses with a given task. In our experiments, we fine-tune the mid-trained models in \S\ref{experiment_mid_training} on synthetically generated samples for the target tasks of RULER~\cite{hsieh2024ruler} SQuADQA and HotpotQA.

During post-training, we apply \ours as a nested learning algorithm. Within each training loop, we first train the model on the prompt with \ours, and then fine-tune the final response to align with the reference. We compare three post-training scenarios. (1) \textit{SFT}: we fine-tune the model directly on the post-training dataset with NTP (no nested learning). (2) \textit{Nested SFT}: we apply nested training strategy with NTP loss. (3) \textit{Nested \ours}: we apply nested training strategy with \ours.

\cref{tab:ruler_qa} shows that post-training with nested \ours (6th rows) outperforms SFT (4th rows) and nested SFT (5th rows). For instance, nested \ours on LaCT-760M improves the average score on SQuADQA by 17\% compared to nested SFT (25.5 vs. 21.8), and by 24.1\% for DeltaNet-1.3B (10.3 vs. 8.3).
These results suggest that NSP provides better task-agnostic learning signals than NTP.

\subsection{Impact of \ours on Test-Time Training (TTT)}
\label{experiment_test_time_training}

\ours can be used during inference to improve performance on a target task. During inference, we apply \ours on the prompt before letting the model generate the final response. In order to maximize memory capacity over long contexts, we provide more direct learning signals by using binary exact match reward ($R^{\text{binary}}$) and a higher RL loss coefficient ($\lambda_{\text{RL}} = 0.4$). We replace \ours with pure SFT on the prompt for comparison.

We apply TTT on the post-trained models for multi-doc QA tasks in \S\ref{experiment_post_training}. The 7th and 8th rows of \cref{tab:ruler_qa} show the results of SFT TTT and \ours TTT, respectively. \ours consistently outperforms SFT during TTT, as in the mid-training and post-training scenarios. Similarly,
\cref{tab:longbench} shows that TTT with \ours yields superior performance compared to SFT across diverse subtasks in LongBench~\cite{bai2024longbench}. This suggests that NSP provides stronger test-time adaptation signals.

\subsection{Analysis}
\label{sec:analysis}

\begin{figure*}[t]
    \centering
    \begin{subfigure}[b]{0.48\textwidth}
        \centering
        \caption{QMSum
        }
        \vspace{-3pt}
         \includegraphics[width=\linewidth]{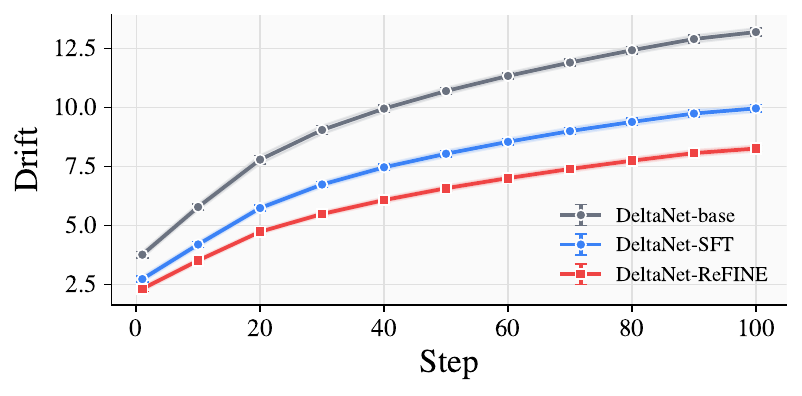}
    \end{subfigure}
    \hfill
    \begin{subfigure}[b]{0.48\textwidth}
        \centering
        \caption{SAMSum}
        \vspace{-3pt}
        \includegraphics[width=\linewidth]{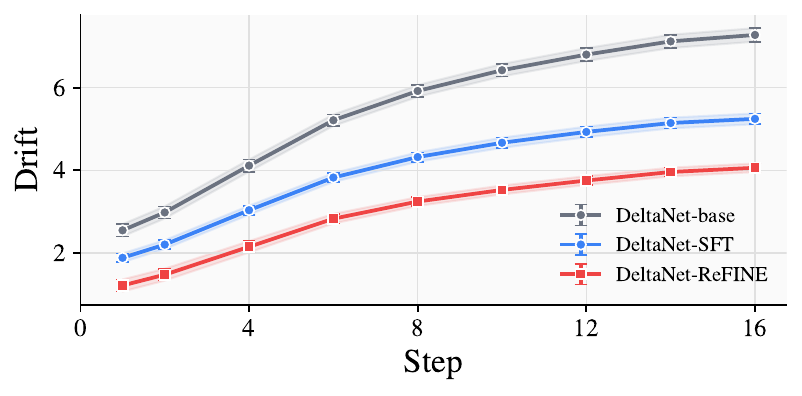}
    \end{subfigure}
    \vspace{-5pt}
    \caption{
        \textbf{Fast Weight Drift.} We measure drift as the Frobenius norm between the fast weight memory matrices induced by predicted and ground truth response trajectories, averaged across layers. Mid-training DeltaNet-1.3B with \ours lowers the drift on QMSum \textbf{(a)} and SAMSum \textbf{(b)} tasks in LongBench~\cite{bai2024longbench} more than SFT. Error bars show $\pm 1\sigma$ across 20 independent generations.
    }
    \vspace{-5pt}
    \label{fig:drift_fw}
\end{figure*}

\paragraph{Analysis of Drift in Fast Weights.} We study how \ours mid-training affects autoregressive drift in fast weight memory. We sample 20 examples from QMSum and SAMSum in LongBench~\cite{bai2024longbench}, and compare pre-trained DeltaNet-1.3B with SFT- and \ours-mid-trained variants. For each model, we generate responses and extract the fast weight memory matrices from all layers after each predicted and ground-truth response token. We measure drift as the Frobenius norm between the memory states induced by the predicted and ground-truth trajectories. As shown in \cref{fig:drift_fw}, \ours consistently reduces drift relative to SFT, suggesting that NSP improves the long-horizon stability of fast weights. More discussion on the drift in fast weights can be found in Appendix~\S\ref{app:additional_analysis}, ~\cref{fig:drift_fw_app,tab:drift_ablation}.

\paragraph{Analysis of Reward Functions.} We analyze the impact of reward functions for \ours. During mid-training, \ours assigns a semantically driven reward to each rollout based on the cosine similarity of the hidden states of predicted and ground truth tokens (\cref{eq:cs_reward}). We replace cosine similarity rewards ($R^{\varphi}$) with binary exact match rewards ($R^{\text{binary}}$), hybrid rewards ($R^{\text{hybrid}}$), or negative Euclidean distance rewards ($R^{\text{NED}}$). $R^{\text{NED}}$ measures the negative Euclidean distance of predicted and ground truth hidden states. \cref{tab:reward_ablation} shows that $R^{\varphi}$ achieves superior performance for mid-training: +1.8\% over $R^{\text{binary}}$ for LaCT-760M and +3.0\% for DeltaNet-1.3B. $R^{\text{NED}}$ performs on par with $R^{\varphi}$ as negative Euclidean distance is monotonic to cosine similarity after advantage standardization. This suggests that similarity-based rewards lead to better generalization. More analysis on TTT can be found in Appendix~\S\ref{app:additional_analysis},~\cref{tab:reward_ablation_ttt}.

\begin{table}[t!]
\centering
\small

\begin{minipage}{0.48\columnwidth}
\centering
\caption{\textbf{Impact of Reward Function on Mid-Training.}
We compare $\mathcal{R}^\text{binary}$, $\mathcal{R}^\text{hybrid}$, $\mathcal{R}^\text{NED}$ and $\mathcal{R}^{\varphi}$ on 12 LongBench tasks.}
\label{tab:reward_ablation}

\footnotesize
\begin{tabular}{l l c}
\toprule
& Reward & Avg. \\
\midrule
\multirow{4}{*}{\shortstack{LaCT-760M\\+\ours MidTr}}
& $\mathcal{R}^{\text{binary}}$ & 16.6 \\
& $\mathcal{R}^{\text{hybrid}}$ & 16.8 \\
& $\mathcal{R}^{\text{NED}}$ &  16.8 \\
& \cellcolor{gray!20} $\mathcal{R}^{\varphi}$ (ours)
& \cellcolor{gray!20} 16.9 \\
\midrule
\multirow{4}{*}{\shortstack{DeltaNet-1.3B\\+\ours MidTr}} 
& $\mathcal{R}^{\text{binary}}$ & 16.5 \\
& $\mathcal{R}^{\text{hybrid}}$ & 16.8 \\
& $\mathcal{R}^{\text{NED}}$ & 17.0 \\
& \cellcolor{gray!20} $\mathcal{R}^{\varphi}$ (ours)
& \cellcolor{gray!20} 17.0 \\
\bottomrule
\end{tabular}
\end{minipage}
\hfill
\begin{minipage}{0.48\columnwidth}
\centering
\caption{\textbf{Impact of Token Selection on Mid-Training.}
We compare uniform, $\arg\max H$, $\arg\min H$, and entropy-weighted sampling.}
\label{tab:es_ablation}
\footnotesize
\begin{tabular}{l l c}
\toprule
& Sampling & Avg. \\
\midrule
\multirow{4}{*}{\shortstack{LaCT-760M\\+\ours MidTr}}
& Uniform  & 16.2 \\
& $\arg\max H$ & 16.4 \\
& $\arg\min H$ & 16.6 \\
\rowcolor{gray!20}
\cellcolor{white}
& Ours & 16.9 \\
\midrule
\multirow{4}{*}{\shortstack{DeltaNet-1.3B\\+\ours MidTr}} 
& Uniform  & 15.9 \\
& $\arg\max H$ & 16.7 \\
& $\arg\min H$ & 16.8 \\
\rowcolor{gray!20}
\cellcolor{white}
& Ours & 17.0 \\
\bottomrule
\end{tabular}
\end{minipage}

\vspace{-5pt}
\end{table}

\paragraph{Analysis of Entropy-Based Token Selection.}
We analyze the role of entropy-based token selection on \ours mid-training. \ours samples a target token from each chunk weighted by the token-level NTP entropy. Rollouts are generated to predict continuations from the sampled target tokens. We repeat the mid-training process after replacing entropy-based sampling with three alternative sampling methods: uniform sampling, maximum entropy selection, and minimum entropy selection. \cref{tab:es_ablation} shows that entropy-weighted sampling achieves the best performance on both models. This suggests that NSP training is most effective when applied to regions with varying levels of uncertainty.

\subsection{Ablations}
\label{sec:ablations}

\begin{wrapfigure}{r}{0.5\columnwidth}
    \vspace{-35pt}
    \centering
    \includegraphics[width=\linewidth]{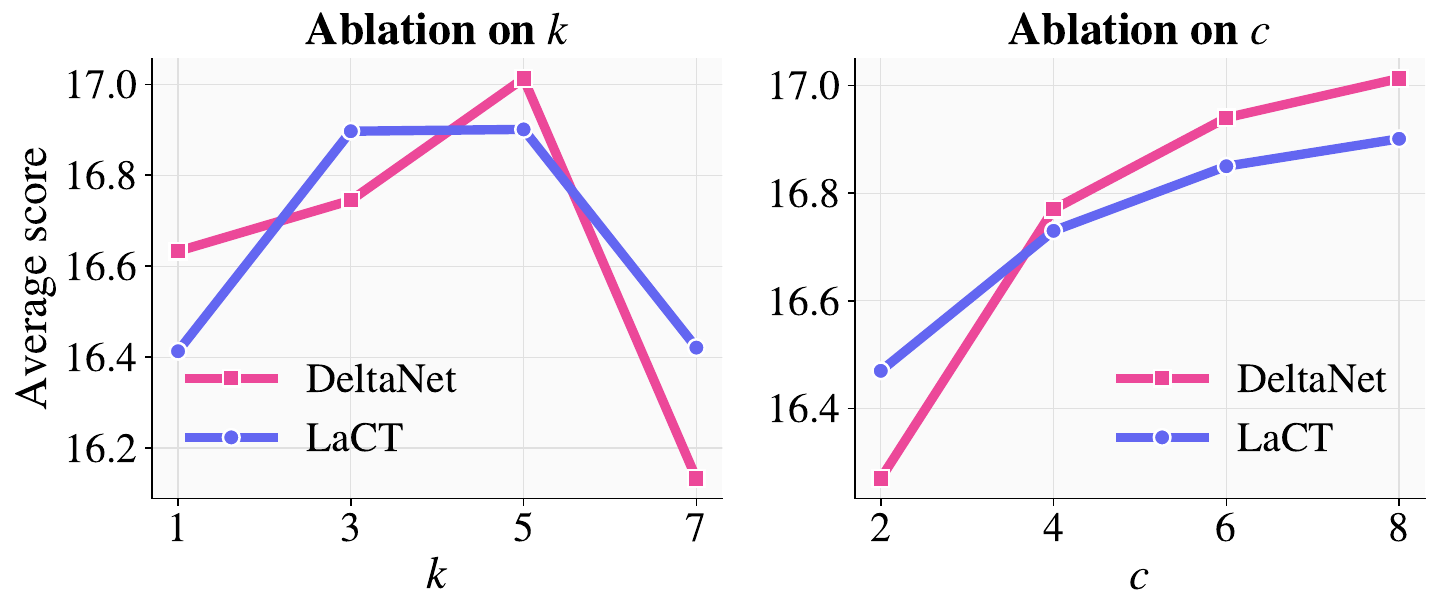}
    \vspace{-20pt}
    \caption{
        \textbf{Ablation on $k$ and $c$.} We mid-train models with different numbers of tokens per rollout $k$ (\textbf{left}) and numbers of chunks per sequence $c$ (\textbf{right}). We evaluate on 12 tasks in LongBench~\cite{bai2024longbench}. With cosine similarity reward, there is an optimal $k$. Higher $c$ leads to more NSP training per sequence, improving overall performance.
    }
    \vspace{-10pt}
    \label{fig:ck_ablation}
\end{wrapfigure}

\paragraph{Rollout Length.} We examine the effect of rollout length $k$ on \ours mid-training, which is the number of tokens to unroll per rollout. $k$ determines how far the model is expected to predict given a prefix. We mid-train both models with different values of $k$ and evaluate on LongBench tasks in \cref{tab:longbench}. We observe that the average score increases until $k=5$ and decreases at $k=7$ (\cref{fig:ck_ablation}, \textit{left}). We hypothesize that the sharpness of the reward starts to degrade when the rollouts extend further. More discussion on the reward distribution can be found in Appendix~\S\ref{app:reward_distribution}.

\paragraph{Number of Chunks per Sequence.} We study the impact of the number of chunks $c$ per sequence on downstream performance by mid-training the models with different numbers of chunks. $c$ determines the number of target tokens sampled based on entropy values, as well as the total number of rollouts per sequence. We evaluate the mid-trained models on LongBench tasks in \cref{tab:longbench}. We find that the average score increases consistently as the number of chunks per sequence increases (\cref{fig:ck_ablation}, \textit{right}). The average score increases from 16.5 ($c=2$) to 16.9 ($c=8$) for LaCT-760M and from 16.3 ($c=2$) to 17.0 ($c=8$) for DeltaNet-1.3B. This indicates that the quality of fast weight initializations increases as the frequency of sequence-level predictions increases.

\section{Discussion}
\label{discussion}

\paragraph{Conclusion.}

We introduce the NSP training objective for fast weight language models to address the limitations of NTP in providing sequence-level feedback. We propose \ours, an RL framework which leverages entropy-based token selection and sequence-level rewards to efficiently train fast weight models under the NSP objective. Our experiments demonstrate that \ours is effective throughout the training lifecycle of fast weight models, showing consistent improvements in long-context benchmarks and reducing inference-time drift. \ours presents RL for NSP as a flexible and practical pathway towards long-context modeling of fast weight architectures.

\paragraph{Limitations.}
First, \ours's cosine similarity reward deteriorates for overly long rollouts, leading to diminishing returns on $k$. Reward functions that capture richer semantic similarity could address this. Second, the optimal rollout length for a given prefix is context-dependent, suggesting that dynamic adjustment may better isolate semantically meaningful regions. 

\paragraph{Future Work.} Fully incorporating NSP into the standard training framework of fast weight models requires architectural changes. Efficient transfer of fast weights across truncated prefixes will significantly accelerate rollout generation, allowing \ours to scale data and compute further.

\section*{Acknowledgements}
All experiments, data collection, and processing activities were conducted at Princeton University, and no Google DeepMind computing resources were used for this work. We thank Yoonsang Lee and Tianyuan Zhang for helpful discussions and valuable insights throughout this project.

\section*{Funding Transparency Statement}

\paragraph{Funding.} Hee Seung Hwang is supported by the Princeton University First-Year Fellowship in the Natural Sciences and Engineering. Sanghyuk Chun is supported by the Francis Robbins Upton Fellowship at Princeton University. Xindi Wu is supported by the Apple Fellowship. 

\paragraph{Competing Interests.} The authors declare no competing interests.

\bibliographystyle{unsrtnat}
\bibliography{references}

\newpage
\clearpage
\appendix
\numberwithin{equation}{section}
\numberwithin{figure}{section}
\numberwithin{table}{section}

\section{Notation}
\label{app:notation}

\begin{table}[h]
\centering
\caption{Glossary and notation.}
\label{tab:notation}
\renewcommand{\arraystretch}{1.4}
\begin{tabularx}{0.85\textwidth}{@{}l X@{}}
\toprule
\textbf{Symbol} & \textbf{Description} \\
\midrule
\multicolumn{2}{@{}l}{\textit{Sequences and Tokens}} \\
$S = (x_1, \ldots, x_T)$ & Input sequence with $T$ tokens \\
$x_t$ & Token at position $t$ \\
$x_{< t}$, $x_{\leq t}$ & Token sequence up to (and including) position $t$ \\
$x_{t+1:t+k}$ & $k$-token ground-truth continuation starting from position $t+1$ \\
$\hat{x}_{t+1:t+k}$ & Generated rollout token sequence from position $t+1$ \\
$T$ & Total sequence length \\
$k$ & Rollout length (number of tokens to predict per rollout) \\
$c$ & Number of chunks per sequence \\
$n$ & Number of rollouts per sampled position \\
$S_i$ & $i$-th chunk of the input sequence \\
$\mathcal{D}$ & Training data distribution \\
\\[-0.5em]
\multicolumn{2}{@{}l}{\textit{Model and Policy}} \\
$\pi_\theta$ & Policy model with parameters $\theta$ \\
$\pi_{\theta_{\mathrm{old}}}$ & Old policy (before update) \\
$p(x_t \mid x_{<t})$ & Conditional probability of token $x_t$ given context \\
$\mathbf{h}^{\text{pred}}(t)$ & Hidden state vector of predicted token at position $t$ \\
$\mathbf{h}^{\text{gt}}(t)$ & Hidden state vector of ground-truth token at position $t$ \\
$\mathbf{h}^{\text{pred}}_k(t)$, $\mathbf{h}^{\text{gt}}_k(t)$ & Matrix of $k$ hidden state vectors starting at position $t+1$ ($\in \mathbb{R}^{k \times d}$) \\
$d$ & Hidden state dimension \\
\\[-0.5em]
\multicolumn{2}{@{}l}{\textit{Entropy and Sampling}} \\
$H_t$ & Entropy at token position $t$ \\
$\tilde{H}_t$ & 1-D average pooled entropy at token position $t$ \\
$\mathcal{T}^* = \{t_1, \ldots, t_c\}$ & Set of sampled high-entropy positions \\
$\mathcal{T}_i$ & Set of token positions in chunk $i$ \\
$p_i(t)$ & Entropy-weighted sampling probability in chunk $i$ for token position $t$ \\
\\[-0.5em]
\multicolumn{2}{@{}l}{\textit{Loss Functions}} \\
$\mathcal{L}_{\mathrm{NTP}}$ & Next-Token Prediction loss (cross-entropy) \\
$\mathcal{L}_{\mathrm{NSP}}$ & Next-Sequence Prediction loss \\
$\mathcal{L}_{\text{seq}}$ & Sequence-level discrepancy measure \\
$\mathcal{J}(\theta)$ & RL objective to maximize \\
\bottomrule
\end{tabularx}
\end{table}

\begin{table}[t!]
\centering
\caption{Glossary and notation (continued).}
\label{tab:notation-continued}
\renewcommand{\arraystretch}{1.4}
\begin{tabularx}{0.85\textwidth}{@{}l X@{}}
\toprule
\textbf{Symbol} & \textbf{Description} \\
\midrule

\multicolumn{2}{@{}l}{\textit{Rewards}} \\
$R(t)$ & Self-supervised reward at position $t$ (shorthand for $R(\hat{x}_{t+1:t+k}, x_{t+1:t+k})$) \\
$R^\varphi_k(t)$ & Cosine similarity reward over $k$-token rollout \\
$R^{\text{binary}}_k(t)$ & Binary exact match reward over $k$-token rollout \\
$R^{\text{hybrid}}_k(t)$ & Hybrid reward ($R^\varphi_k + R^{\text{binary}}_k$) \\
$R^{\text{NED}}_k(t)$ & Negative Euclidean distance reward over $k$-token rollout \\
$\mathcal{R}_{\mathcal{T}^*}$ & Set of rewards for all sampled positions in $\mathcal{T}^*$ \\
$\cos(\cdot, \cdot)$ & Cosine similarity function \\
$\mathcal{I}[\cdot]$ & Indicator function \\
$o_{\mathcal{T}^*}$ & Set of all rollouts at sampled positions \\
\\[-0.5em]
\multicolumn{2}{@{}l}{\textit{Fast Weight Architecture}} \\
$\mathbf{W}_t$ & Fast weight matrix at token position $t$ \\
$\eta$ & Learning rate for fast weight update \\
$\mathbf{k}_t, \mathbf{v}_t, \mathbf{q}_t, \mathbf{o}_t, $ & Key, value, query, and output vector representations at position $t$ \\

\bottomrule
\end{tabularx}
\end{table}

\newpage

\section{Ethics Statement}
\label{app:ethics}

We believe that the broader societal impact of our work aligns with the consequences of improving long-context modeling capabilities of fast weight language models. We adopt verified and publicly available datasets and classes of models for our experiments, and therefore do not see a clear negative impact on the research community and beyond. However, we advise users to take caution in the downstream applications of our work.

\section{Related Work}
\label{app:related}

\paragraph{Multi-Token Prediction.} Predicting more than one token as a training objective has been explored in previous studies.~\citet{gloeckle2024better} tackle this problem by estimating $k$ tokens in parallel using $k$ independent output heads. This approach achieves significant gains in throughput by applying architectural modifications to the language model, with minimal degradation of performance on downstream tasks. However, this approach is limited in capturing dependencies among the predicted tokens and relies on a fixed prediction horizon $k$. \ours computes rewards based on each multi-token rollout as a whole, capturing the semantic connection among them. 

\citet{liu2025sequential} use diffusion-based generation to predict multiple masked tokens simultaneously and optimize the cross-entropy loss between the predicted tokens and masked ground truth tokens. Their method is primarily designed for standard attention-based transformer architectures that predict the next token with masking. However, \ours is designed to train fast weight language models that store information and update their parameters with a fundamentally different set of rules. We view this work as valuable motivation, although it is evaluated on a different class of models.

\paragraph{Continued Pre-Training with RL.} Studies have explored using RL as a tool for the next-token prediction objective.~\citet{dong2025reinforcement} sample reasoning traces before next-token prediction and assign rewards based on the similarity of the byte-sequences of predicted and ground truth tokens.~\citet{hatamizadeh2025rlp} take a similar approach by sampling reasoning traces for next-token prediction but assigning rewards by measuring the gap between the log-likelihood of the ground truth token with and without the reasoning trace as context. However, these works focus on attention-based transformer models (DeepSeek-R1-Distill-Qwen-14B~\cite{guo2025deepseek} and Qwen3-1.7B-Base~\cite{qwen3technicalreport}, respectively) and assume basic reasoning capabilities that enable exploration with Chain-of-Thought. Whether RL can be used for pre-trained fast weight models without prior instruction tuning or human preference optimization, especially in long-context settings, is yet to be explored. Further, \ours leverages RL for training on the NSP objective which optimizes sequence-level predictions rather than single-token predictions under the NTP objective.

\paragraph{TTT in Language Models.} Recent work has explored training standard transformer-based language models during inference to improve performance on target tasks. These methods usually involve extracting task-related learning signals from the model itself for offline adaptation. \citet{akyurek2024surprising} generate relevant in-context examples from the given task and train the model on those examples before generating the final answer to the actual task. RL-based methods extract pseudo-labels by aggregating multiple responses to the same task and assigning rewards to each response based on their similarity to the pseudo-label~\cite{zuo2025ttrl, huang2024self}. Recently, context-based test-time training has also been explored in transformer-based language models. In an attempt to overcome the inherent limitations of static attention, \citet{bansal2025let} execute gradient updates on the query projection matrices using the context while keeping other parameters frozen. These approaches apply task-aware or task-agnostic TTT methods to standard transformer-based language models. TTT with \ours, on the other hand, aims to improve contextual adaptation and memory of fast weights for long contexts, which is a novel setting that has not yet been explored.  

\paragraph{Efficient Attention Variants.} Standard transformer-based language models with full attention incur quadratic computational complexity as a function of context length. Recent work has developed techniques to reduce the computational and memory overhead in long-context modeling. Sparse attention addresses this problem by introducing computational sparsity in the original attention mechanism. Grouped Query Attention~\cite{ainslie2023gqa} employs sparsity along the head dimension to assign each query head to different groups that share a single key-value head. Sliding-window attention~\cite{child2019generating, beltagy2020longformer, zaheer2020big} architectures leverage sparsity along the context by computing local attention on a fixed number of contiguous tokens. 

Linear attention approximates the softmax kernel in the attention formulation to achieve linear computational complexity. Linformer~\cite{wang2020linformer} replaces the attention mechanism with low-rank matrix operations, which has been shown to be effective for sequence processing but limited in autoregressive generation. Performer~\cite{choromanski2020rethinking} uses orthogonal random features to approximate softmax kernels in the attention, achieving linear complexity without employing low rank matrices. Similarly, Linear Transformer~\cite{katharopoulos2020transformers} approximates the softmax with linear dot-product of kernel feature maps. The success of linear attention has motivated the development of architectures that operate on linear computational complexity by design, including State Space Models (SSMs), such as Mamba~\cite{gu2024mamba, dao2024transformers}. Linear attention can also be read as a recurrent model: its key-value outer products accumulate into a fixed-size state that is updated once per token, so decoding requires constant memory regardless of context length. A growing body of work builds on this view, which we group into three families below: gated linear recurrent models, fast weight models, and hybrid recurrent attention models.

\paragraph{Gated Linear Recurrent Models.} This family augments linear attention with learned, input-dependent gates that control how the recurrent state decays and accumulates information over time. RetNet~\cite{sun2023retentive} introduces an exponentially decayed retention state with equivalent parallel, recurrent, and chunkwise recurrent forms. RWKV~\cite{peng2023rwkv} combines receptance-controlled key and value aggregation with parallelizable training and recurrent inference. RWKV-7~\cite{peng2025rwkv} introduces a generalized delta rule update with vector-valued gates and input-dependent learning rates, improving the expressivity and state tracking capabilities of the recurrent state. Gated Linear Attention~\cite{yang2024gla} introduces data-dependent decay over a matrix-valued linear attention state, while HGRN2~\cite{qin2024hgrn2} expands the recurrent state to improve its representational capacity. Similarly, the matrix LSTM component of xLSTM~\cite{beck2024xlstm} maintains a normalized matrix memory using exponential input and forget gates. These models primarily improve recurrent language modeling through more expressive gating, forgetting, and state expansion mechanisms. \ours is agnostic to this choice of gating or state design: rather than proposing a new recurrent architecture, it targets how the recurrent update of an existing architecture is trained so that it stays accurate once the model starts conditioning on its own predictions.

\paragraph{Fast Weight Models.} Fast weight models~\cite{schlag2021linear} cast the recurrent state itself as a set of fast weights, parameters that are written to and read from at every step through key-value operations. DeltaNet~\cite{yang2024parallelizing} replaces the additive update of linear attention with a delta rule that corrects the value currently associated with an incoming key before writing a new value. Gated DeltaNet~\cite{yang2024gated} combines this corrective update with input-dependent forgetting, Kimi Delta Attention~\cite{team2025kimi} extends this decay mechanism with channel-wise decay for finer-grained control over memory retention, and Gated DeltaNet-2~\cite{hatamizadeh2026gated} further decouples the erase and write gates, allowing the model to control removal of old associations and insertion of new values independently. Other work derives fast weight updates more explicitly from online learning objectives. Longhorn~\cite{liu2025longhorn} formulates its state update as an amortized solution to online associative recall. Test-Time Training (TTT) layers~\cite{sun2024learning} treat the recurrent state as a predictive model updated through self-supervised learning. Titans~\cite{behrouz2024titans} introduces a neural long-term memory that learns to memorize the preceding context at test time and combines this persistent memory with attention-based short-term memory. MesaNet~\cite{vonoswald2026mesanet} updates its memory using a locally optimal test-time regression rule. These works design or refine the form of the online update rule itself. Our \ours instead leaves the update rule fixed and changes how the model is trained to use it, so that the memory produced during training matches the one the model actually relies on during autoregressive generation.

\paragraph{Hybrid Recurrent Attention Models.} This family augments a recurrent backbone with a limited amount of explicit attention to compensate for the fixed-size state's bounded capacity to compress the entire preceding context. Griffin~\cite{de2024griffin} interleaves gated recurrent layers with local attention, and RecurrentGemma~\cite{botev2024recurrentgemma} scales this architecture to pretrained and instruction-tuned language models. Jamba~\cite{lieber2024jamba} similarly interleaves Mamba and Transformer layers, while BASED~\cite{arora2024simple} combines linear attention with sliding-window attention to balance recurrent state efficiency against associative recall.

Across these families, contextual information is compressed into an input-dependent state rather than retained in an ever-growing key-value cache. Prior work has primarily focused on designing more expressive state representations and update rules. In contrast, \ours does not introduce a new recurrent architecture. It provides a training framework for fast weight models that improves how their existing online memory updates are learned and used through sequence-level reinforcement learning.

\section{Datasets and Benchmarks}
\label{app:datasets}

We summarize the datasets and benchmarks used for training and evaluation in \cref{tab:datasets}. The Long-Data-Collections~\cite{longdatacollections} (Apache 2.0 License) dataset contains a 68.8B-token pre-training corpus subsampled from RedPajama~\cite{weber2024redpajama}, Pile~\cite{gao2020pile}, UL2 Oscar~\cite{tay2022ul2}, NI~\cite{mishra2022cross}, and P3~\cite{bach2022promptsource}. We use a 200M-token subset of Long-Data-Collections for mid-training.
For LongBench~\cite{bai2024longbench} (MIT License), we select 12 English-language subtasks. We leave out MuSiQue and GovReport tasks because they have fewer than 20 samples under 16K tokens. For RULER~\cite{hsieh2024ruler} (Apache 2.0 license) multi-doc QA tasks, we use the official synthetic data generation script to create evaluation samples.

\begin{table}[h]
\centering
\caption{Summary of datasets and benchmarks used across training phases.}
\label{tab:datasets}
\small
\begin{tabular}{l l l l l}
\toprule
\textbf{Phase} & \textbf{Dataset} &  Metric  & \textbf{Context} & \textbf{Size} \\
\midrule
\multirow{3}{*}{Mid-training} & Long-Data-Collections & - & 16K & $\sim$200M tokens \\
& RULER NIAH & recall & 4K/8K/16K & 500 per context \\
& Booksum & NTP Accuracy, CE loss & $\leq$16K & 9600 \\
\midrule
\multirow{2}{*}{Post-training} & RULER SQuADQA &  recall & 4K/8K/16K & 1600 train / 200 test \\
& RULER HotpotQA & recall & 4K/8K/16K & 1600 train / 200 test \\
\midrule
\multirow{6}{*}{Test-time} & NarrativeQA (NQ) & F1 &  $\leq$16K & 56 \\ 
& Qasper (QR) & F1 & $\leq$16K & 184 \\
& MultiFieldQA (MF) & F1 & $\leq$16K & 136	 \\
& HotpotQA (HP) & F1 & $\leq$16K & 96 \\
& 2WikiMHQA (2W) & F1  & $\leq$16K & 184 \\
& QMSum (QM) & rouge  & $\leq$16K & 104 \\
& MultiNews (MN) & rouge  & $\leq$16K & 192  \\
& SAMSum (SS) & rouge  & $\leq$16K & 152 \\
& TREC (TC) & accuracy  & $\leq$16K & 200  \\
& TriviaQA (TQ) & accuracy  & $\leq$16K & 120  \\
& LCC (LC) & code similarity & $\leq$16K & 488 \\
& RepoBench-P (RP) & code similarity & $\leq$16K & 320  \\
\midrule
\multirow{9}{*}{Commonsense} & PIQA & accuracy  &  & All  \\
& HellaSwag (Hella.) & normalized accuracy  &  & All  \\
& WinoGrande (Wino.) & accuracy  &  & All  \\
& ARC-e &  accuracy &  & All  \\ 
& ARC-c & normalized accuracy  &  & All \\
& Wikitext (Wiki.) & perplexity  &  & All  \\
& LAMBADA (LMB.) & perplexity, accuracy  &  & All  \\ 
& FDA & recall &  & All  \\
& SWDE & recall  &  & All  \\

\bottomrule
\end{tabular}
\vspace{-10pt}
\end{table}

\section{Training Configuration}
\label{app:configuration}
\begin{table}[h]
\centering
\caption{Training hyperparameters.}
\label{tab:hyperparameters}
\small
\renewcommand{\arraystretch}{1.15}
\begin{tabular}{@{}lc@{\hskip 1.5em}lc@{}}
\toprule
\textbf{Parameter} & \textbf{Value} & \textbf{Parameter} & \textbf{Value} \\
\midrule
Actor gradient clip                 & 0.2   & Learning rate                & $10^{-6}$ \\
Mid-Train batch size                & 128   & Adam $(\beta_1,\beta_2)$     & $(0.9,\, 0.999)$ \\
Post-Train batch size               & 64    & Weight decay                 & 0.01 \\
Test-time Train batch size          & 8     & Sampling temperature $\tau$  & 1.0 \\
\addlinespace[0.5em]
Mid-Train PPO mini-batch size       & 32    & Max prompt length            & 16{,}384 \\
Post-Train PPO mini-batch size      & 16    & Entropy loss coefficient     & 0 \\
Test-time Train PPO mini-batch size & 4     & KL loss coefficient          & 0 \\
\addlinespace[0.5em]
Mid-Train $\lambda_{\text{RL}}$     & 0.2   & $n$ (rollouts / position)    & 1 \\
Post-Train $\lambda_{\text{RL}}$    & 0.2   & $k$ (rollout length)         & 5 \\
Test-time Train $\lambda_{\text{RL}}$ & 0.4 & $c$ (chunks / sequence)      & 8 \\
\addlinespace[0.5em]
Mid-Train $\lambda_{\text{SFT}}$    & 1.0   & Mid-Train reward             & $R^\varphi$ \\
Post-Train $\lambda_{\text{SFT}}$   & 1.0   & Post-Train reward            & $R^{\text{hybrid}}$ \\
Test-time Train $\lambda_{\text{SFT}}$ & 1.0 & Test-time Train reward      & $R^{\text{binary}}$ \\
\bottomrule
\end{tabular}
\vspace{-10pt}
\end{table}

\paragraph{Hyperparameters.} We provide the training hyperparameters used for \ours during all training phases in \cref{tab:hyperparameters}. We only adjust the train batch size, reward function, and RL loss coefficient across training phases, while keeping all else equal. 

\paragraph{Compute.} We use 8 L40 GPUs for mid-training and post-training, and 4 L40 GPUs for TTT. We use fewer GPUs for TTT because the train batch size for TTT is smaller (8 samples per batch) compared to other training phases (128 for mid-training, 64 for post-training). Mid-training LaCT-760M and DeltaNet-1.3B with \ours on 200M tokens from Long-Data-Collections~\cite{longdatacollections} at 16K context takes approximately 24 hours.

\section{Evaluation Details}
\label{app:evaluation}

We evaluate models on RULER~\cite{hsieh2024ruler} NIAH tasks with Language Model Evaluation Harness~\cite{eval-harness}. For evaluation on RULER QA and LongBench~\cite{bai2024longbench} tasks, we implement custom pipelines in order to be compatible with the test-time training framework. 
We follow the official metrics and configurations from Language Model Evaluation Harness~\cite{eval-harness} in our implementations. 

\section{Additional Analysis}
\label{app:additional_analysis}

\paragraph{Additional Baselines.} We compare \ours against baseline methods that aim to provide sequence-level loss for standard transformer models. Motivated by \citet{fang2024wrong}, we experiment with a variant of pure SFT, where the CE loss of $k$ tokens at high-entropy positions sampled as in \cref{sec:framework} are up-weighted by a factor of two (we denote this method as SFT*). We also implement architectural multi-token prediction method motivated by \citet{gloeckle2024better}. During training, we clone the last layer into $k=4$ heads (we choose $k=4$ because it divides 16384, our training context length) and train the model with normal CE loss. Each head is designed to predict a single token within a contiguous block of $k=4$ tokens. During inference, we discard all prediction heads except one and generate tokens autoregressively (we denote this method as MTP). 
\Cref{tab:ruler_niah_baseline,tab:ruler_qa_baseline,tab:longbench_baseline} show the performance gain of training the models with baseline methods during mid-, post-, and test-time training, as well as the original comparison of \ours and SFT. The results show that the maximum performance gain achievable is highest for \ours, indicating that \ours's RL approach in providing sequence-level learning signal is most effective. 

\begin{table*}[t!]
    \centering    
    \footnotesize
    \caption{
    \textbf{Performance on Long-Context Retrieval Tasks with Additional Baselines.} 
    We evaluate mid-trained (\textbf{MidTr}) models on the NIAH tasks in RULER at 4K, 8K, and 16K context lengths. Highest scores in each category are highlighted in \textbf{bold}.
    } 
    \resizebox{\textwidth}{!}{
    \begin{tabular}{c c | c c c c | c c c c | c c c c | c c c c}
\toprule
&& \multicolumn{4}{c|}{RULER S-NIAH} 
& \multicolumn{4}{c|}{RULER MK-NIAH} 
& \multicolumn{4}{c|}{RULER MQ-NIAH} 
& \multicolumn{4}{c}{RULER MV-NIAH} \\

Model & MidTr
& 4K
& 8K
& 16K
& Avg
& 4K
& 8K
& 16K
& Avg 
& 4K
& 8K
& 16K
& Avg
& 4K
& 8K
& 16K
& Avg \\
            
\midrule

\multirow{5}{*}{\shortstack{LaCT\\760M}}
& -
& 98.8
& 91.2
& 95.8
& 95.3
& 70.2
& 44.8
& 24.2
& 46.2
& 39.2
& 24.9
& 17.4
& 27.1
& 40.6
& 27.0
& 17.7
& 28.4 \\

& SFT
& 98.4
& 90.8
& 97.6
& 95.6
& \textbf{70.6}
& 45.0
& 24.2
& 46.6
& \textbf{40.7}
& 25.2
& 17.4
& 27.8
& \textbf{41.7}
& 26.3
& 17.7
& 28.5 \\

& \cellcolor{gray!20}\ours
& \cellcolor{gray!20}\textbf{99.0}
& \cellcolor{gray!20}93.0
& \cellcolor{gray!20}96.8
& \cellcolor{gray!20}96.3
& \cellcolor{gray!20}70.4
& \cellcolor{gray!20}\textbf{46.0}
& \cellcolor{gray!20}\textbf{26.6}
& \cellcolor{gray!20}\textbf{47.7}
& \cellcolor{gray!20}40.5
& \cellcolor{gray!20}\textbf{25.8}
& \cellcolor{gray!20}\textbf{18.0}
& \cellcolor{gray!20}\textbf{28.1}
& \cellcolor{gray!20}41.2
& \cellcolor{gray!20}\textbf{29.5}
& \cellcolor{gray!20}\textbf{18.6}
& \cellcolor{gray!20}\textbf{29.8} \\

\cmidrule(lr){2-18}

& SFT*
& 98.8	
& 92.6
& \textbf{98.4}
& 96.6
& 69.4
& 44.8
& 24.2
& 46.1
& 41.2
& 25.5
& 17.2
& 28.0
& 42.0
& 27.1
& 17.8
& 29.0 \\

& MTP
& 98.6	
& \textbf{95.0}
& 98.0
& \textbf{97.2}
& 70.0
& 46.8
& 24.0
& 46.9
& 43.6
& 25.4
& 17.1
& 28.7
& 41.9
& 26.0
& 17.9
& 28.6 \\
            
\midrule

\multirow{5}{*}{\shortstack{DeltaNet\\1.3B}}
& -
& \textbf{100.0}
& \textbf{100.0}
& \textbf{100.0}
& \textbf{100.0}
& 23.6
& 17.8
& 3.4
& 14.9
& 27.7
& 3.9
& 1.8
& 11.1
& 23.9
& 5.4
& 2.0
& 10.4 \\

& SFT
& \textbf{100.0}
& \textbf{100.0}
& \textbf{100.0}
& \textbf{100.0}
& 23.8
& 19.2
& 7.8
& 16.9
& 33.2
& 16.8
& 3.5
& 17.8
& 28.3
& 19.3
& 3.7
& 17.1 \\

& \cellcolor{gray!20}\ours
& \cellcolor{gray!20}\textbf{100.0}
& \cellcolor{gray!20}\textbf{100.0}
& \cellcolor{gray!20}99.8
& \cellcolor{gray!20}99.9
& \cellcolor{gray!20}\textbf{25.0}
& \cellcolor{gray!20}\textbf{21.4}
& \cellcolor{gray!20}\textbf{8.8}
& \cellcolor{gray!20}\textbf{18.4}
& \cellcolor{gray!20}\textbf{37.3}
& \cellcolor{gray!20}\textbf{18.8}
& \cellcolor{gray!20}3.6
& \cellcolor{gray!20}\textbf{19.9}
& \cellcolor{gray!20}\textbf{29.0}
& \cellcolor{gray!20}\textbf{20.4}
& \cellcolor{gray!20}\textbf{4.0}
& \cellcolor{gray!20}\textbf{17.8} \\

\cmidrule(lr){2-18}

& SFT*
& \textbf{100.0}
& \textbf{100.0}
& 99.8
& 99.9	
& 24.2
& 20.6
& 6.2
& 17.0
& 32.9
& 15.7
& \textbf{4.4}
& 17.7
& 28.5
& 19.4
& 3.4
& 17.1 \\

& MTP
& \textbf{100.0}	
& \textbf{100.0}
& \textbf{100.0}
& \textbf{100.0}
& 24.4
& 17.2
& 2.6
& 14.7
& 29.2
& 2.3
& 1.9
& 11.1
& 24.3
& 2.4
& 1.7
& 9.5 \\

\bottomrule
\end{tabular}
}
\label{tab:ruler_niah_baseline}
\end{table*}

\begin{table*}[t]
    \centering
    \scriptsize
    \caption{\textbf{Performance on Multi-Doc QA Tasks with Additional Baselines.} We evaluate various training strategies during mid-training (\textbf{MidTr}), post-training (\textbf{PostTr}), and test-time training (\textbf{TTT}) for multi-document question and answering tasks. First and second highest scores on each task are highlighted in \textbf{bold} and \underline{underline}, respectively.
    }
            \begin{tabular}{cccc | c c c c | c c c c }
\toprule
&&&& \multicolumn{4}{c|}{RULER SQuADQA} 
& \multicolumn{4}{c}{RULER HotpotQA} \\
& MidTr & PostTr & TTT
& 4K
& 8K
& 16K
& Avg
& 4K
& 8K
& 16K
& Avg \\ 
\midrule
\multirow{10}{*}{\shortstack{LaCT\\760M}}
& - & - & -
& 17.5	
& 14.0
& 6.5
& 12.7
& 20.5
& 14.5
& 12.0
& 15.7 \\

& SFT & - & -
& 19.5	
& 14.0
& 6.0
& 13.2
& 21.5
& 15.0
& 12.0
& 16.2 \\

\rowcolor{gray!20}
\cellcolor{white}
& \ours & - & -
& 20.5	
& 15.5
& 8.0
& 14.7
& 23.0
& 16.0
& 12.5
& 17.2 \\

& \ours & SFT & -
& 30.5	
& 20.0
& 7.5
& 19.3
& 25.0
& 16.5
& 11.5
& 17.7 \\

& \ours & Nested SFT & -
& 38.0	
& 20.0
& 7.5
& 21.8
& 24.0
& 16.0
& 12.5
& 17.5 \\

& \cellcolor{gray!20}\ours & \cellcolor{gray!20}Nested \ours & \cellcolor{gray!20}-
& \cellcolor{gray!20}\underline{43.5}	
& \cellcolor{gray!20}\underline{24.5}
& \cellcolor{gray!20}8.5
& \cellcolor{gray!20}\underline{25.5}
& \cellcolor{gray!20}\underline{27.0}
& \cellcolor{gray!20}19.5
& \cellcolor{gray!20}13.0
& \cellcolor{gray!20}19.8 \\

& \ours & Nested \ours & SFT
& \underline{43.5}
& 21.5
& 8.5
& 24.5
& 26.5
& \underline{24.0}
& 13.0
& \underline{21.2} \\

& \cellcolor{gray!20}\ours & \cellcolor{gray!20}Nested \ours & \cellcolor{gray!20}\ours
& \cellcolor{gray!20}\textbf{45.5}	
& \cellcolor{gray!20}\textbf{25.5}
& \cellcolor{gray!20}\underline{10.0}
& \cellcolor{gray!20}\textbf{27.0}
& \cellcolor{gray!20}\textbf{28.5}
& \cellcolor{gray!20}\textbf{25.5}
& \cellcolor{gray!20}\textbf{15.0}
& \cellcolor{gray!20}\textbf{23.0} \\

\cmidrule(lr){2-12}

& SFT* & - & -
& 22.0	
& 13.0	
& 6.5	
& 13.8
& 20.0	
& 13.5	
& 13.0
& 15.5 \\

& SFT* & Nested SFT* & -
& 35.5	
& 23.5		
& \textbf{11.0}	
& 23.3	
& 24.5	
& 19.5	
& \underline{14.5}
& 19.5	 \\

& SFT* & Nested SFT* & SFT*
& 35.5		
& 24.5	
& 8.0	
& 22.7	
& 25.5	
& 21.5	
& 14.0	
& 20.3 \\

& MTP & - & -
& 20.5	
& 13.0	
& 7.0	
& 13.5	
& 20.0	
& 17.0	
& 9.0	
& 15.3 \\

& MTP & Nested MTP & -
& 34.0	
& 23.0	
& 8.0			
& 21.7	
& 25.5	
& 19.0	
& 12.5	
& 19.0 \\

& MTP & Nested MTP & MTP
& 33.5						
& 20.5
& 9.5
& 21.2
& 24.5
& 22.0
& 13.0	
& 19.8	 \\

\midrule
\multirow{10}{*}{\shortstack{DeltaNet\\1.3B}}
& - & - & -
& 11.0	
& 5.0
& 2.0
& 6.0
& 10.0
& 3.0
& 2.5
& 5.2		 \\

& SFT & - & -
& 9.0	
& 6.0
& 3.0
& 6.0
& 9.0
& 8.0
& 4.5
& 7.2	  \\

\rowcolor{gray!20}
\cellcolor{white}
& \ours & - & -
& 10.0	
& 6.5
& 4.0
& 6.8
& 12.0
& 9.5
& 5.5
& 9.0	  \\
    
& \ours & SFT & -
& 11.0	
& 7.0
& 5.0
& 7.7
& 14.0
& 11.0
& 6.5
& 10.5	\\

& \ours & Nested SFT & -
& 12.0	
& 8.0
& 5.0
& 8.3
& \underline{16.5}
& 10.0
& 7.5
& 11.3 \\

& \cellcolor{gray!20}\ours & \cellcolor{gray!20}Nested \ours & \cellcolor{gray!20}-
& \cellcolor{gray!20}14.0	
& \cellcolor{gray!20}\underline{10.5}
& \cellcolor{gray!20}\underline{6.5}
& \cellcolor{gray!20}10.3
& \cellcolor{gray!20}15.0
& \cellcolor{gray!20}11.0
& \cellcolor{gray!20}\textbf{8.5}
& \cellcolor{gray!20}11.5 \\

& \ours & Nested \ours & SFT
& \underline{16.5}
& \underline{10.5}
& \underline{6.5}
& \underline{11.2}
& 16.0
& 10.5
& 6.5
& 11.0 \\

\rowcolor{gray!20}
\cellcolor{white}
& \ours & Nested \ours & \ours
& \textbf{17.5}
& \textbf{12.5}
& \textbf{7.0}
& \textbf{12.3}
& \textbf{18.0}
& \textbf{13.5}
& \underline{8.0}
& \textbf{13.2} \\

\cmidrule(lr){2-12}

& SFT* & - & -
& 9.0	
& 5.5	
& 3.5
& 6.0	
& 12.5	
& 7.5	
& 4.5	
& 8.2 \\

& SFT* & Nested SFT* & -
& 13.5	
& 8.5	
& 5.5	
& 9.2	
& 12.5	
& 8.5	
& 6.5	
& 9.2 \\

& SFT* & Nested SFT* & SFT*
& 15.0	
& 7.5	
& 6.0	
& 9.5	
& 15.5	 	
& 10.5	
& 9.0	
& \underline{11.7}	 	 \\

& MTP & - & -
& 9.5	
& 4.0	
& 2.0		
& 5.2
& 9.5	
& 5.5	
& 3.0	
& 6.0 \\

& MTP & Nested MTP & -
& 12.5	
& 7.0		
& 4.0	
& 7.8	
& 11.5	
& 9.5	
& 6.5	
& 9.2 \\

& MTP & Nested MTP & MTP
& 12.0	
& 5.0	
& 3.5	
& 6.8
& 15.5	
& \underline{12.0}
& 7.0	
& 11.5 \\

\bottomrule
\end{tabular}
\vspace{-5pt}
\label{tab:ruler_qa_baseline}
\end{table*}

\begin{table*}[t]
    \centering
    \small
    \caption{\textbf{Performance on Long-Context Tasks in LongBench with Additional Baselines.} We study the impact of the learning algorithm during mid-training (\textbf{MidTr}) and test-time training (\textbf{TTT}) on tasks with long-context. \textbf{SFT} denotes the supervised fine-tuning with next-token prediction.
    We evaluate on 12 tasks in LongBench, filtered for samples with at most 16K tokens.
    }
    \resizebox{\textwidth}{!}{
        \begin{tabular}{c | c c | c c c | c c | c c | c c c | c c | c}
            \toprule
             &&
             & \multicolumn{3}{c|}{Single-doc QA} 
             & \multicolumn{2}{c|}{Multi-doc QA} 
             & \multicolumn{2}{c|}{Summarization}
             & \multicolumn{3}{c|}{Few-shot QA}
             & \multicolumn{2}{c|}{Coding}
             & \\
& MidTr & TTT 
& NQ
& QR
& MF
& HP
& 2W
& QM
& MN
& SS
& TC
& TQ
& LC
& RP
& Avg \\
\midrule
\multirow{7}{*}{\shortstack{LaCT\\760M}}
& - & -
& 6.5
& 10.5
& 7.2
& 11.7
& 9.8
& 13.6	
& 9.2	
& 14.2
& 10.5
& 8.0
& 26.7
& 29.7
& 13.1 \\

& SFT & - 
& 5.8
& 10.1
& 7.4
& 12.6
& 9.2
& 13.1	
& 10.5	
& 13.8
& 7.5
& 12.2
& 29.8
& 30.1
& 13.5 \\

& \cellcolor{gray!20}\ours & \cellcolor{gray!20}-
& \cellcolor{gray!20}6.5
& \cellcolor{gray!20}11.1	
& \cellcolor{gray!20}12.6
& \cellcolor{gray!20}\textbf{19.6}	
& \cellcolor{gray!20}18.0	
& \cellcolor{gray!20}\textbf{14.3}
& \cellcolor{gray!20}\underline{15.9}
& \cellcolor{gray!20}\underline{17.0}
& \cellcolor{gray!20}11.0
& \cellcolor{gray!20}12.9
& \cellcolor{gray!20}\textbf{32.9}
& \cellcolor{gray!20}31.1
& \cellcolor{gray!20}16.9 \\

&\ours & SFT
& 5.1	
& \underline{12.6}
& \underline{13.5}
& 15.9
& \underline{18.4}
& 13.2
& \textbf{16.2}
& \textbf{17.4}
& 12.5
& \textbf{16.2}
& 32.0
& \underline{31.4}
& \underline{17.0} \\

& \cellcolor{gray!20}\ours & \cellcolor{gray!20}\ours
& \cellcolor{gray!20}\underline{6.7}
& \cellcolor{gray!20}\textbf{14.5}	
& \cellcolor{gray!20}\textbf{14.1}	
& \cellcolor{gray!20}\underline{18.4}
& \cellcolor{gray!20}\textbf{22.8}	
& \cellcolor{gray!20}13.9
& \cellcolor{gray!20}\underline{15.9}
& \cellcolor{gray!20}\textbf{17.4}	
& \cellcolor{gray!20}\textbf{15.5}	
& \cellcolor{gray!20}11.8	
& \cellcolor{gray!20}\underline{32.2}
& \cellcolor{gray!20}\textbf{32.3}	
& \cellcolor{gray!20}\textbf{18.0} \\

\cmidrule(lr){2-16}
& SFT* & - 
& 4.8	
& 11.0	
& 8.0
& 18.1
& 9.5
& 13.4
& 11.2
& 14.1
& 11.0
& 12.2
& \underline{28.0}
& 30.5
& 14.3  \\

& SFT* & SFT*
& \textbf{7.9}
& 10.2	
& 8.0	
& 16.8	
& 14.0	
& \underline{14.1}
& 12.7	
& 14.9	
& 13.0	
& 13.8
& 30.0	
& 30.2	
& 15.5 \\

& MTP & -
& 4.5	
& 9.0	
& 6.4	
& 17.0	
& 6.3
& 13.0
& 11.1
& 13.1
& \underline{15.0}
& 13.3
& 28.4
& 29.4
& 13.9 \\

& MTP & MTP
& 4.2	
& 9.1	
& 7.0
& 12.8	
& 10.1		
& 13.1	
& 11.8	
& 15.6	
& \underline{15.0}
& \underline{14.5}
& 26.1
& 30.2	
& 14.1	  \\

\midrule
\multirow{7}{*}{\shortstack{DeltaNet\\1.3B}}
& - & -
& 6.5
& 8.7	
& 10.0	
& 4.8	
& 6.4	
& 12.4	
& \underline{16.3}
& 9.4	
& 17.7	
& 15.1	
& 33.8	
& 29.0 
& 14.2 \\

& SFT & -
& 5.7	
& 9.4	
& 8.3	
& 9.2
& 8.6
& 14.7	
& 16.1	
& 15.5	
& 22.9	
& 15.2	
& 33.1	
& 29.2 
& 15.7 \\

& \cellcolor{gray!20}\ours & \cellcolor{gray!20}-
& \cellcolor{gray!20}6.5
& \cellcolor{gray!20}9.5
& \cellcolor{gray!20}10.1	
& \cellcolor{gray!20}\textbf{9.6}
& \cellcolor{gray!20}8.7
& \cellcolor{gray!20}\textbf{15.2}	
& \cellcolor{gray!20}15.0	
& \cellcolor{gray!20}16.2	
& \cellcolor{gray!20}25.0	
& \cellcolor{gray!20}\underline{21.4}	
& \cellcolor{gray!20}\textbf{35.9}	
& \cellcolor{gray!20}\textbf{31.1}	
& \cellcolor{gray!20}\underline{17.0} \\

& \ours & SFT
& 7.2
& \underline{9.6}
& 10.5
& 6.8
& 7.2
& \underline{14.9}
& 16.2
& \textbf{17.3}
& \underline{28.0}
& 16.6
& 34.1
& 29.2
& 16.5 \\

& \cellcolor{gray!20}\ours & \cellcolor{gray!20}\ours
& \cellcolor{gray!20}\textbf{7.5}	
& \cellcolor{gray!20}9.2	
& \cellcolor{gray!20}\underline{11.5}	
& \cellcolor{gray!20}\underline{9.5}
& \cellcolor{gray!20}8.6
& \cellcolor{gray!20}14.7	
& \cellcolor{gray!20}16.1	
& \cellcolor{gray!20}\underline{16.5}
& \cellcolor{gray!20}\textbf{31.5}	
& \cellcolor{gray!20}\textbf{24.7}	
& \cellcolor{gray!20}\underline{35.2}
& \cellcolor{gray!20}30.0
& \cellcolor{gray!20}\textbf{17.9} \\

\cmidrule(lr){2-16}
& SFT* & - 
& 5.2	
& 9.4
& 10.4
& 6.5
& \underline{9.2}
& 14.5
& \textbf{16.7}
& 14.8
& 23.0
& 21.1
& 32.3
& 28.1
& 15.9 \\

& SFT* & SFT*
& 5.9	
& \textbf{9.9}
& \textbf{13.0}
& 5.1	
& \textbf{11.0}	
& 14.1	
& 16.2	
& 15.7	
& \underline{28.0}
& 20.6	
& 33.8	
& \underline{30.5}
& \underline{17.0}   \\

& MTP & -
& 5.8
& 7.5
& 8.7
& 8.5
& 8.7
& 13.7	
& 16.0
& 11.3
& 18.5
& 15.9
& 32.1
& 27.7
& 14.5 \\

& MTP & MTP
& \underline{6.7}
& 7.7	
& 10.2			
& 5.7	
& 8.1	
& 13.7	
& 15.9	
& 10.8	
& 18.0	
& 18.2	
& 31.8	
& 28.3	
& 14.6  \\

\bottomrule
\end{tabular}
}
\vspace{-10pt}
\label{tab:longbench_baseline}
\end{table*}

\paragraph{Compute Matched SFT Baselines.}

\begin{table*}[t]
    \centering
    \small
    \caption{\textbf{Compute-Matched Comparison of Performance on LongBench Tasks.} 
    We study the impact of the learning algorithm during mid-training (\textbf{MidTr}) on long-context tasks. 
    \textbf{CM-SFT} denotes compute-matched supervised fine-tuning with next-token prediction.
    We evaluate on 12 tasks in LongBench, filtered for samples with at most 16K tokens.
    Details are similar to \cref{tab:ruler_qa}.
    }
    \resizebox{\textwidth}{!}{
        \begin{tabular}{c c | c c c | c c | c c | c c c | c c | c}
            \toprule
             &&
             \multicolumn{3}{c|}{Single-doc QA} 
             & \multicolumn{2}{c|}{Multi-doc QA} 
             & \multicolumn{2}{c|}{Summarization}
             & \multicolumn{3}{c|}{Few-shot QA}
             & \multicolumn{2}{c|}{Coding}
             & \\
& MidTr
& NQ
& QR
& MF
& HP
& 2W
& QM
& MN
& SS
& TC
& TQ
& LC
& RP
& Avg \\
\midrule
\multirow{3}{*}{\shortstack{LaCT\\760M}}
& -
& 6.5
& 10.5
& 7.2
& 11.7
& 9.8
& 13.6	
& 9.2	
& 14.2
& 10.5
& 8.0
& 26.7
& 29.7
& 13.1 \\

& SFT (2500 steps)
& \textbf{6.9}
& 9.5
& 9.1
& 15.2
& 11.4
& 14.1
& 8.3
& \textbf{17.0}
& \textbf{12.0}
& \textbf{13.3}
& 27.9
& \textbf{31.5}
& 14.7 \\

& \cellcolor{gray!20}\ours (100 steps)
& \cellcolor{gray!20}6.5
& \cellcolor{gray!20}\textbf{11.1}	
& \cellcolor{gray!20}\textbf{12.6}
& \cellcolor{gray!20}\textbf{19.6}	
& \cellcolor{gray!20}\textbf{18.0}	
& \cellcolor{gray!20}\textbf{14.3}
& \cellcolor{gray!20}\textbf{15.9}
& \cellcolor{gray!20}\textbf{17.0}
& \cellcolor{gray!20}11.0
& \cellcolor{gray!20}12.9
& \cellcolor{gray!20}\textbf{32.9}
& \cellcolor{gray!20}31.1
& \cellcolor{gray!20}\textbf{16.9} \\

\midrule
\multirow{3}{*}{\shortstack{DeltaNet\\1.3B}}
& -
& \textbf{6.5}
& 8.7	
& 10.0	
& 4.8	
& 6.4	
& 12.4	
& 16.3	
& 9.4	
& 17.7	
& 15.1	
& 33.8	
& 29.0
& 14.2 \\

& SFT (1400 steps)
& 5.3	
& 8.7	
& \textbf{11.5}	
& 8.6
& \textbf{9.2}	
& 14.7	
& \textbf{17.3}	
& \textbf{17.7}	
& \textbf{27.5}	
& 20.1	
& 32.3	
& 28.1 
& 16.8 \\

& \cellcolor{gray!20}\ours (100 steps)
& \cellcolor{gray!20}\textbf{6.5}
& \cellcolor{gray!20}\textbf{9.5}
& \cellcolor{gray!20}10.1	
& \cellcolor{gray!20}\textbf{9.6}
& \cellcolor{gray!20}8.7
& \cellcolor{gray!20}\textbf{15.2}	
& \cellcolor{gray!20}15.0	
& \cellcolor{gray!20}16.2	
& \cellcolor{gray!20}25.0	
& \cellcolor{gray!20}\textbf{21.4}	
& \cellcolor{gray!20}\textbf{35.9}	
& \cellcolor{gray!20}\textbf{31.1}	
& \cellcolor{gray!20}\textbf{17.0} \\

\bottomrule
\end{tabular}
}
\vspace{-10pt}
\label{tab:longbench_cm}
\end{table*}

Each training step in \ours takes longer than SFT (11x for DeltaNet-1.3B and 23x for LaCT-760M). The vLLM~\cite{kwon2023efficient} engine in veRL~\cite{sheng2024hybridflow} does not yet support fast weight models for rollout generation. The computational overhead should reduce once comparable infrastructure is available for fast weight models.

We experiment with a compute-matched SFT baseline and compare its LongBench~\cite{bai2024longbench} performance with \ours. We mid-train DeltaNet-1.3B for 1400 steps on 14x more data, and LaCT-760M for 2500 steps on 25x more data (both more generous than the 11x and 23x compute ratios) while fixing \ours mid-training at 100 steps. \cref{tab:longbench_cm} shows that mid-training with \ours still outperforms compute-matched SFT baselines, indicating that our method is more efficient after accounting for compute. 

\paragraph{Scaling Experiments.}

In order to demonstrate that \ours can be applied to larger models, we experiment with DeltaNet-2.7B, which is twice as big as DeltaNet-1.3B. Larger pre-trained LaCT models are not yet publicly available. \cref{tab:longbench_scaling} shows that \ours mid-training still outperforms SFT on LongBench, showing that our method can scale to larger model parameters.

\begin{table}[t]
    \centering
    \small
    \caption{\textbf{Scaling Mid-Training to DeltaNet 2.7B.}
    We compare SFT and \ours mid-training on DeltaNet 2.7B using LongBench tasks.
    Higher is better. Best scores are highlighted in \textbf{bold}.
    }
    \resizebox{\columnwidth}{!}{
        \begin{tabular}{c c | c c c | c c | c c | c c c | c c | c}
            \toprule
             &&
             \multicolumn{3}{c|}{Single-doc QA} 
             & \multicolumn{2}{c|}{Multi-doc QA} 
             & \multicolumn{2}{c|}{Summarization}
             & \multicolumn{3}{c|}{Few-shot QA}
             & \multicolumn{2}{c|}{Coding}
             & \\
& MidTr
& NQ
& QR
& MF
& HP
& 2W
& QM
& MN
& SS
& TC
& TQ
& LC
& RP
& Avg \\
\midrule
\multirow{2}{*}{\shortstack{DeltaNet\\2.7B}}
& SFT
& 3.8
& 9.7
& \textbf{11.6}
& 6.6
& \textbf{13.2}
& \textbf{14.2}
& 17.0
& 8.2
& 28.5
& 14.9
& \textbf{36.5}
& \textbf{31.0}
& 16.3 \\

& \cellcolor{gray!20}\ours
& \cellcolor{gray!20}\textbf{8.6}
& \cellcolor{gray!20}\textbf{9.9}
& \cellcolor{gray!20}\textbf{11.6}
& \cellcolor{gray!20}\textbf{6.8}
& \cellcolor{gray!20}10.4
& \cellcolor{gray!20}14.0
& \cellcolor{gray!20}\textbf{17.8}
& \cellcolor{gray!20}\textbf{14.0}
& \cellcolor{gray!20}\textbf{36.5}
& \cellcolor{gray!20}\textbf{18.5}
& \cellcolor{gray!20}35.8
& \cellcolor{gray!20}28.9
& \cellcolor{gray!20}\textbf{17.7} \\

\bottomrule
\end{tabular}
}
\vspace{-8pt}
\label{tab:longbench_scaling}
\end{table}

\begin{figure*}[t!]
    \centering
    \includegraphics[width=\textwidth]{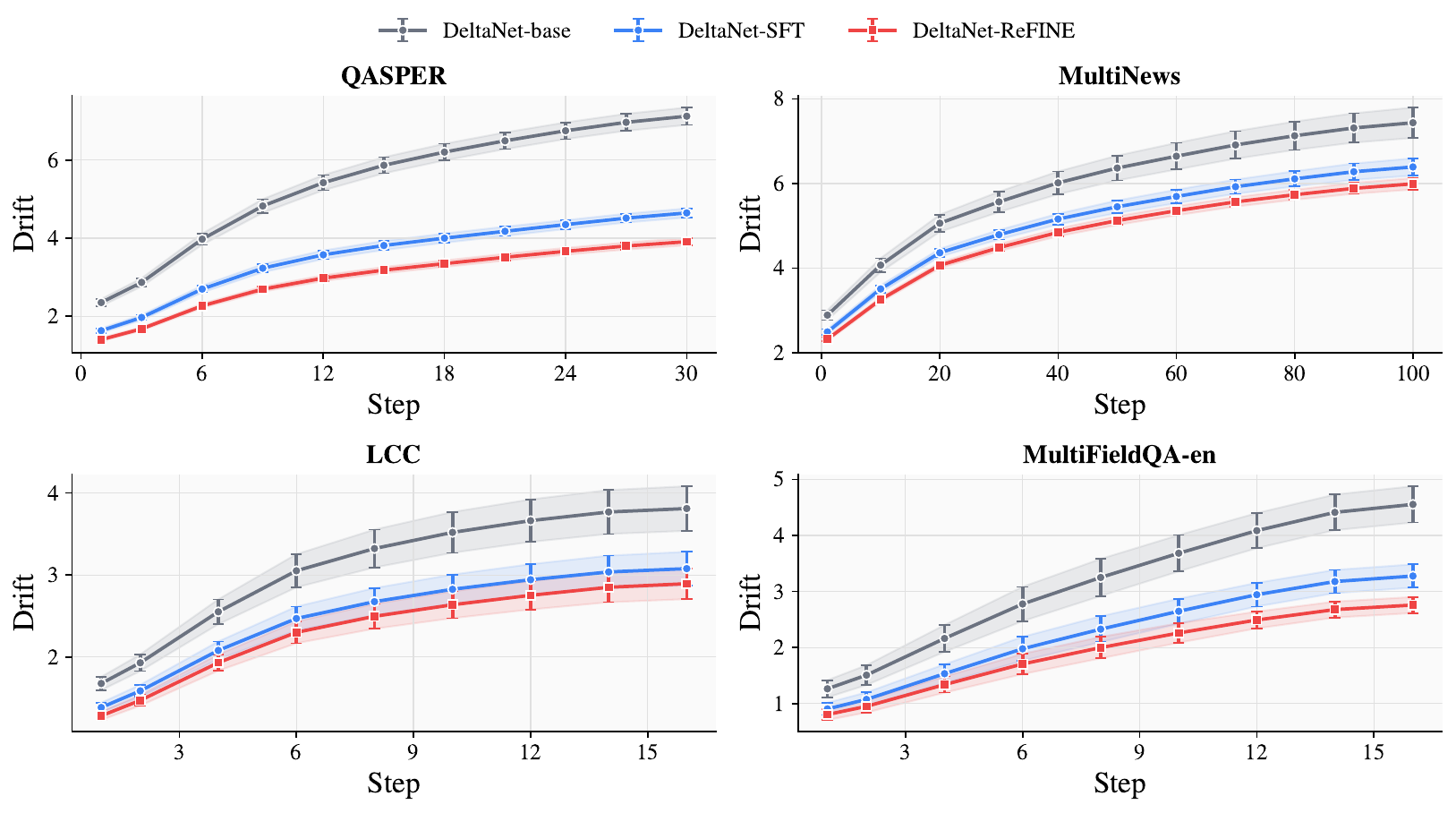}
    \vspace{-15pt}
    \caption{ \textbf{Additional Analysis on Drift in Fast Weights.} We measure drift as the Frobenius norm between the fast weight memory matrices induced by predicted and ground truth response trajectories, averaged across layers. Mid-training DeltaNet-1.3B with \ours lowers the drift on various tasks in LongBench~\cite{bai2024longbench} more than SFT. Error bars show $\pm 1\sigma$ across 20 independent generations.
    }
    \vspace{-15pt}
    \label{fig:drift_fw_app}
\end{figure*}

\paragraph{Additional Analysis of Drift in Fast Weights.} We repeat the fast weight drift analysis in \cref{sec:analysis} on four additional tasks in LongBench~\cite{bai2024longbench}. We take 20 random samples from each task and compute the Frobenius norm of the difference between the fast weight matrices for predicted and ground truth response trajectories, averaged across the layers. ~\cref{fig:drift_fw_app} shows that mid-training DeltaNet-1.3B with \ours reduces drift in fast weights across tasks compared to SFT, suggesting that NSP is beneficial for long-horizon memory stability.  

\begin{table}[t]
\centering
\small
\caption{\textbf{Fast Weight Drift across LongBench Tasks.}
We report the mean  $\pm1 \sigma$ of drift measured by the Frobenius norm between fast weight memory states induced by predicted and ground-truth trajectories. Lower is better. The lowest values are \textbf{bolded}.}
\resizebox{\linewidth}{!}{
\begin{tabular}{l|c|cc|cccccc}
\toprule
Model & MidTr & $k$ & $c$ & QMSum & SAMSum & QASPER & MultiNews & LCC & MultiFieldQA-en \\
\midrule
\multirow{4}{*}{\shortstack{DeltaNet\\1.3B}}
& \cellcolor{gray!20} \ours 
& \cellcolor{gray!20} 5 
& \cellcolor{gray!20} 8 
& \cellcolor{gray!20} $\mathbf{8.322 \pm 0.135}$ 
& \cellcolor{gray!20} $\mathbf{4.134 \pm 0.112}$ 
& \cellcolor{gray!20} $\mathbf{4.024 \pm 0.094}$ 
& \cellcolor{gray!20} $6.145 \pm 0.136$ 
& \cellcolor{gray!20} $\mathbf{2.938 \pm 0.191}$ 
& \cellcolor{gray!20} $\mathbf{2.828 \pm 0.146}$ \\
& \ours & 3 & 8 & $8.416 \pm 0.133$ & $4.171 \pm 0.104$ & $4.053 \pm 0.096$ & $6.085 \pm 0.139$ & $2.963 \pm 0.195$ & $2.886 \pm 0.159$ \\
& \ours & 5 & 4 & $8.446 \pm 0.136$ & $4.186 \pm 0.117$ & $4.048 \pm 0.096$ & $\mathbf{6.076 \pm 0.135}$ & $2.952 \pm 0.193$ & $2.858 \pm 0.162$ \\
& \ours & 5 & 2 & $8.560 \pm 0.133$ & $4.233 \pm 0.106$ & $4.088 \pm 0.096$ & $6.177 \pm 0.153$ & $2.958 \pm 0.193$ & $2.898 \pm 0.155$ \\
\bottomrule
\end{tabular}
}
\label{tab:drift_ablation}
\end{table}

We further study how the rollout length $k$ and the number of chunks $c$ affect fast weight drift. 
As shown in \cref{tab:drift_ablation}, reducing the rollout length from $k=5$ to $k=3$ increases drift across most LongBench tasks. 
Similarly, reducing the number of chunks from $c=8$ to $c=4$ or $c=2$ generally leads to higher drift. 
These trends align with the ablations in \cref{sec:ablations}, where configurations with higher drift also achieve lower average LongBench~\cite{bai2024longbench} performance than the default setting of $k=5$ and $c=8$. 
Together, these results support the hypothesis that NSP improves downstream performance partly by stabilizing recurrent fast weight trajectories over long horizons.

\paragraph{Entropy Distribution.}
\label{app:entropy_distribution}

\begin{figure*}[h!]
    \centering
    \begin{subfigure}[b]{0.48\textwidth}
        \centering
        \caption{LaCT-760M}
        \vspace{-5pt}
        \begin{tikzpicture}
            \node (img1) {\includegraphics[width=\linewidth]{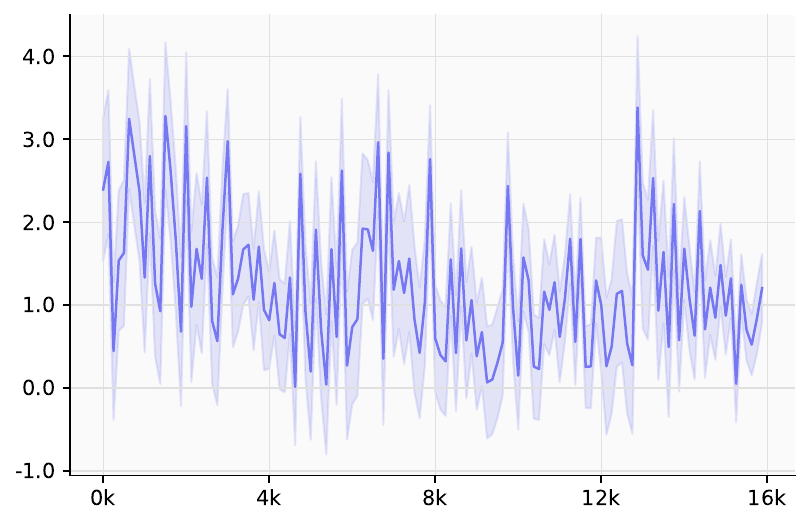}};
            \node[left=of img1, rotate=90, anchor=center, xshift=0.7cm, yshift=-0.9cm, overlay] {\footnotesize{Entropy}};
            \node[below=of img1, node distance=0cm, yshift=1.1cm, overlay] {\footnotesize{Token Index}};
        \end{tikzpicture}
        \vspace{-8pt}
        \label{fig:entropy_lact}
    \end{subfigure}
    \hfill
    \begin{subfigure}[b]{0.48\textwidth}
        \centering
        \caption{DeltaNet-1.3B}
        \vspace{-5pt}
        \begin{tikzpicture}
            \node (img2) {\includegraphics[width=\linewidth]{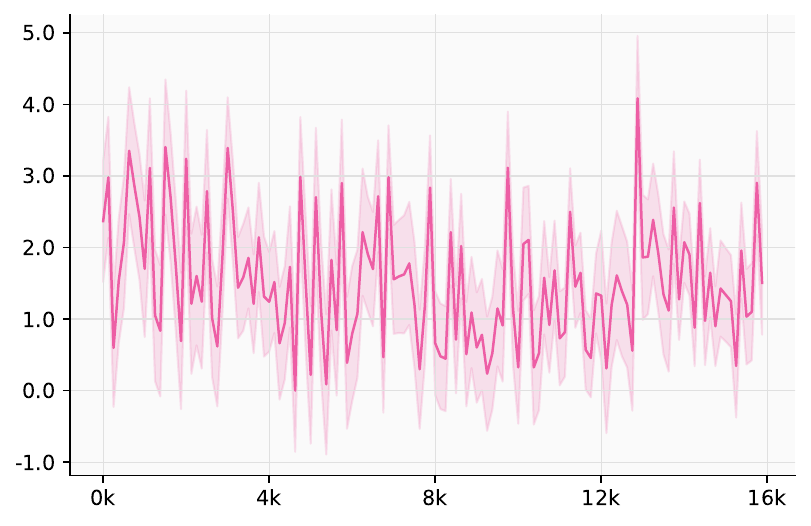}};
            \node[left=of img2, rotate=90, anchor=center, xshift=0.7cm, yshift=-0.9cm, overlay] {\footnotesize{Entropy}};
            \node[below=of img2, node distance=0cm, yshift=1.1cm, overlay] {\footnotesize{Token Index}};
        \end{tikzpicture}
        \vspace{-8pt}
        \label{fig:entropy_deltanet}
    \end{subfigure}
    \vspace{+5pt}
    \caption{\textbf{NTP Entropy Distribution.} We compute the token-level NTP entropy of a randomly selected sample using LaCT-760M \textbf{(a)} and DeltaNet-1.3B \textbf{(b)}. We use the same sample to extract the entropy distribution from both models.
    }
    \label{fig:entropy}
\end{figure*}

We report the NTP entropy distribution of a randomly selected sample in order to illustrate the effects of entropy-based token selection in \ours (\cref{fig:entropy}). We find that there are no index-dependent patterns in the distribution, which justifies per-chunk target token sampling weighted by entropy.

\paragraph{Reward Distribution.}
\label{app:reward_distribution}

\begin{figure*}[h!]
    \centering
    \vspace{-10pt}
    \begin{subfigure}[b]{0.48\textwidth}
        \centering
        \caption{LaCT-760M (Mean)}
        \vspace{-5pt}
        \begin{tikzpicture}
            \node (img1) {\includegraphics[width=\linewidth]{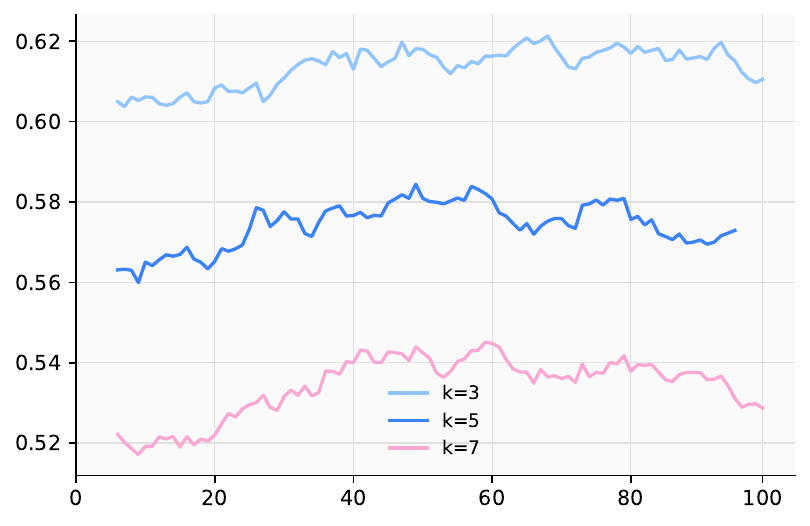}};
            \node[left=of img1, rotate=90, anchor=center, xshift=0.3cm, yshift=-0.9cm] {\footnotesize{Reward Mean}};
            \node[below=of img1, node distance=0cm, yshift=1.1cm] {\footnotesize{Steps}};
        \end{tikzpicture}
        \label{fig:reward_mean_lact}
    \end{subfigure}
    \hfill
    \begin{subfigure}[b]{0.48\textwidth}
        \centering
        \caption{DeltaNet-1.3B (Mean)}
        \vspace{-5pt}
        \begin{tikzpicture}
            \node (img2) {\includegraphics[width=\linewidth]{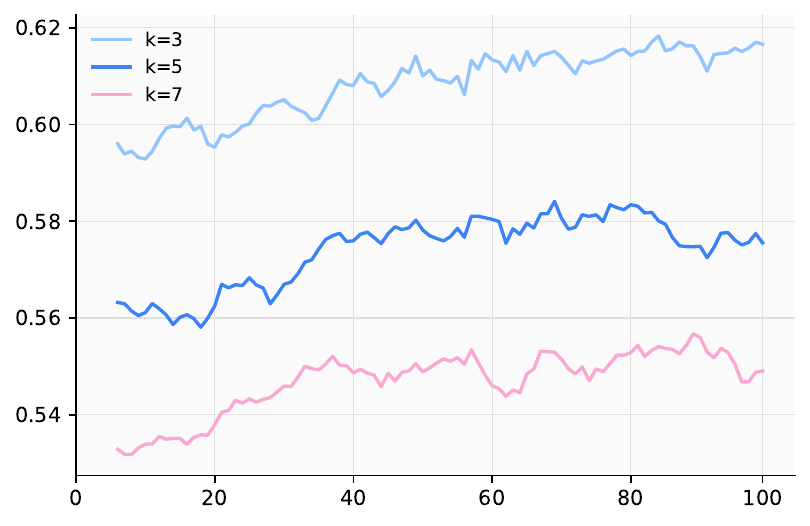}};
            \node[left=of img2, rotate=90, anchor=center, xshift=0.3cm, yshift=-0.9cm] {\footnotesize{Reward Mean}};
            \node[below=of img2, node distance=0cm, yshift=1.1cm] {\footnotesize{Steps}};
        \end{tikzpicture}
        \label{fig:reward_mean_deltanet}
    \end{subfigure}

    \vspace{-5pt}

    \begin{subfigure}[b]{0.48\textwidth}
        \centering
        \caption{LaCT-760M (Std)}
        \vspace{-5pt}
        \begin{tikzpicture}
            \node (img3) {\includegraphics[width=\linewidth]{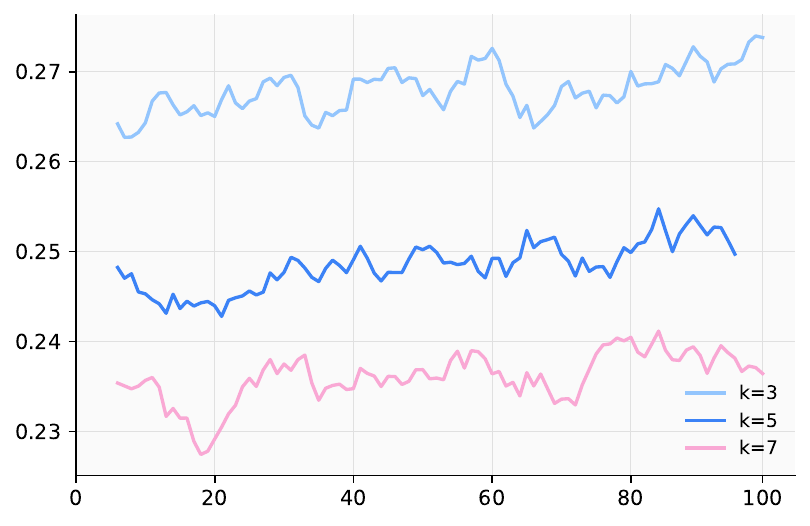}};
            \node[left=of img3, rotate=90, anchor=center, xshift=0.3cm, yshift=-0.9cm] {\footnotesize{Reward Std}};
            \node[below=of img3, node distance=0cm, yshift=1.1cm] {\footnotesize{Steps}};
        \end{tikzpicture}
        \label{fig:reward_std_lact}
    \end{subfigure}
    \hfill
    \begin{subfigure}[b]{0.48\textwidth}
        \centering
        \caption{DeltaNet-1.3B (Std)}
        \vspace{-5pt}
        \begin{tikzpicture}
            \node (img4) {\includegraphics[width=\linewidth]{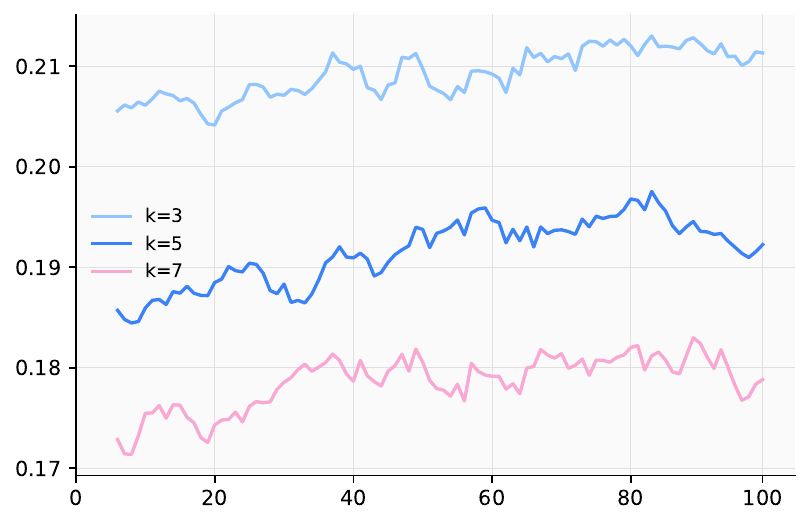}};
            \node[left=of img4, rotate=90, anchor=center, xshift=0.3cm, yshift=-0.9cm] {\footnotesize{Reward Std}};
            \node[below=of img4, node distance=0cm, yshift=1.1cm] {\footnotesize{Steps}};
        \end{tikzpicture}
        \label{fig:reward_std_deltanet}
    \end{subfigure}
    \vspace{-20pt}
    \caption{\textbf{Reward Distribution.} We report the mean and standard deviation of the cosine similarity reward during mid-training for different values of $k$. As the rollout length increases, the mean reward (\textbf{a}, \textbf{b}) decreases and the standard deviation (\textbf{c}, \textbf{d}) also decreases.
    }
    \label{fig:reward}
    \vspace{-5pt}
\end{figure*}

In order to investigate the stability of mid-training with \ours, we track the cosine similarity reward across training steps for different rollout lengths ($k=3,5,7$), as shown in \cref{fig:reward}. We find that the reward distribution remains stable throughout training. However, as rollout length increases, both the mean and variance of the reward decrease, suggesting that the learning signal may lose sharpness for larger $k$.

\begin{table*}[th]
\scriptsize
\centering
\caption{\textbf{Performance on Short-Context Tasks.} 
We evaluate mid-trained models on short-context benchmarks to verify that \ours does not cause catastrophic forgetting.} 
\begin{tabular}{l | c c | c c | c c c c c c | c }
\toprule
Models
& Wiki 
& LMB.
& FDA
& SWDE
& LMB.
& PIQA
& Hella.
& Wino.
& ARC-e
& ARC-c
& Avg \\

& ppl $\downarrow$
& ppl $\downarrow$
& recall $\uparrow$
& recall $\uparrow$
& acc $\uparrow$
& acc $\uparrow$
& acc\_n $\uparrow$
& acc $\uparrow$
& acc $\uparrow$
& acc\_n $\uparrow$
&  \\
\midrule
LaCT-760M 
& 20.8
& 29.8
& 36.7
& 66.0
& 32.4
& 67.5
& 41.5
& 53.4
& 48.4
& 27.6
& 45.1 \\

+SFT MidTr
& 20.7
& 28.9
& 37.2
& 66.2
& 32.9
& 67.4
& 41.4
& 52.5
& 48.4
& 27.1
& 45.0 \\

+\ours MidTr
& 20.8
& 30.2
& 36.1
& 66.5
& 32.3
& 67.3
& 41.4
& 52.8
& 48.8
& 27.5
& 45.0 \\
                  
\midrule
DeltaNet-1.3B
& 16.7
& 14.7
& 18.0
& 54.4
& 43.0
& 70.8
& 50.5
& 53.7
& 58.3
& 25.9
& 50.4 \\

+SFT MidTr
& 16.8
& 15.3
& 17.7
& 54.5
& 42.7
& 71.1
& 50.1
& 53.8
& 58.4
& 26.4
& 50.4 \\

+\ours MidTr
& 16.8
& 15.6
& 17.5
& 54.5
& 42.2
& 71.1
& 50.2
& 53.6
& 58.1
& 26.2
& 50.2 \\
         
\bottomrule
\end{tabular}
\label{tab:icl_cr}
\end{table*}

\paragraph{Performance on Short-Context Tasks.} We investigate whether enhancing long-context handling capabilities of fast weight models leads to degradation on out-of-distribution tasks such as short-context in-context retrieval and commonsense reasoning. We evaluate mid-trained models on 9 relevant short-context tasks with lm-evaluation-harness~\cite{eval-harness}:  PIQA~\cite{bisk2020piqa}, HellaSwag (Hella.)~\cite{zellers2019hellaswag},
WinoGrande (Wino.)~\cite{sakaguchi2020winogrande}, ARC-easy (ARC-e), ARC-challenge (ARC-c)~\cite{clark2018think}, Wikitext (Wiki.)~\cite{merity2016pointer}, LAMBADA (LMB.)~\cite{paperno2016lambada}, FDA~\cite{arora2023language}, and SWDE~\cite{lockard2019openceres}. \cref{tab:icl_cr} shows that \ours sustains performance in these tasks, suggesting that NSP complements NTP without inducing catastrophic forgetting~\cite{van2024continual}.

\begin{table}[t!]
\centering
\small
\caption{\textbf{Impact of reward function on TTT.}
We compare different TTT reward strategies on 12 tasks in LongBench: binary exact match between predicted and ground truth completion ($\mathcal{R}^{\text{binary}}$), and cosine similarity of the hidden states of predicted and ground truth completions ($\mathcal{R}^{\varphi}$).
}
\small
\begin{tabular}{l l l l l c}
\toprule
& MidTr   &  TTT &  MidTr Reward & TTT Reward & Avg. Score \\
\midrule
\multirow{4}{*}{\shortstack{LaCT-760M}}
& \ours & - & $\mathcal{R}^{\varphi}$  & -   & 16.9 \\
& \ours & SFT & $\mathcal{R}^{\varphi}$  & -   & 17.0 \\
& \ours & \ours & $\mathcal{R}^{\varphi}$  & $\mathcal{R}^{\varphi}$   & 17.5 \\
& \cellcolor{gray!20}\ours 
& \cellcolor{gray!20}\ours 
& \cellcolor{gray!20} $\mathcal{R}^{\varphi}$
& \cellcolor{gray!20} $\mathcal{R}^{\text{binary}}$
& \cellcolor{gray!20} 18.0 \\
\midrule
\multirow{4}{*}{\shortstack{DeltaNet-1.3B}} 
& \ours & - & $\mathcal{R}^{\varphi}$  & -   & 17.0 \\
& \ours & SFT & $\mathcal{R}^{\varphi}$  & -   & 16.5 \\
& \ours & \ours & $\mathcal{R}^{\varphi}$  & $\mathcal{R}^{\varphi}$  &  17.6 \\
& \cellcolor{gray!20}\ours 
& \cellcolor{gray!20}\ours
& \cellcolor{gray!20} $\mathcal{R}^{\varphi}$
& \cellcolor{gray!20} $\mathcal{R}^{\text{binary}}$
& \cellcolor{gray!20} 17.9 \\
\bottomrule
\end{tabular}
\vspace{-10pt}

\label{tab:reward_ablation_ttt}

\end{table}

\paragraph{Impact of Reward Functions on TTT.} \ours uses binary exact match reward $\mathcal{R}^{\text{binary}}$ during test-time for offline adaptation of the model before generating the response. We repeat TTT with cosine similarity reward instead and report the average score on LongBench tasks (\cref{tab:longbench}). \Cref{tab:reward_ablation_ttt} shows that the binary reward is optimal for TTT, but the cosine similarity reward also performs well, higher than pure SFT for TTT.

\clearpage
\section{Qualitative Examples} 
\label{app:qualitative}

We provide qualitative examples of cosine similarity reward assignment during mid-training in \cref{ex:rewards}.
We randomly sample prefixes from the mid-training dataset and generate four $k=5$ continuations each using the pre-trained LaCT-760M model. The cosine similarity reward is designed to capture the semantic similarity between the predicted and ground truth continuations. The highest reward values for each example are highlighted in bold. The examples demonstrate that the reward effectively captures semantic similarity beyond exact lexical matching. 
For instance, in example 2, ``enjoyed every minute of it'' achieves near-perfect alignment (0.961) with ``loved every minute of it'', while semantically divergent predictions like ``would not recommend it at'' receive much lower scores (0.463). Similarly, example 6 shows sensitivity to mathematical concepts, where ``is a convergent integral'' receives a high score (0.758) for preserving the convergence concept from ``is also convergent according'', while less relevant predictions such as ``shall not be less than'' receive lower rewards (0.512). These examples illustrate that the cosine similarity reward provides meaningful learning signals for training the model to generate semantically coherent continuations.

\begin{table}[h!]
\centering
\small
\caption{Qualitative examples of cosine similarity reward assignment during mid-training. GT denotes the ground truth continuation, and P1--P4 denote the four predicted continuations generated by the model.}
\begin{tabular}{@{}c l >{\raggedright\arraybackslash}p{6cm} r@{}}
\toprule
\rowcolor{gray!15}
\textbf{Example} & \textbf{Type} & \textbf{Text} & \textbf{Cosine Reward} \\
\midrule
\multirow{5}{*}{\textbf{Ex. 1}} 
& \cellcolor{blue!10}\textbf{GT} & \cellcolor{blue!10}\emph{very laid back lodge} & \cellcolor{blue!10}-- \\
& P1 & perfect hill            & 0.508 \\
& P2 & unique hotel to spend a & 0.498 \\
& P3 & great place to stay     & \cellcolor{yellow!25}\textbf{0.535} \\
& P4 & top-notch l             & 0.418 \\
\cmidrule(lr){1-4}

\multirow{5}{*}{\textbf{Ex. 2}} 
& \cellcolor{blue!10}\textbf{GT} & \cellcolor{blue!10}\emph{loved every minute of it} & \cellcolor{blue!10}-- \\
& P1 & would not recommend it at & 0.463 \\
& P2 & were so impressed by this & 0.566 \\
& P3 & enjoyed every minute of it & \cellcolor{yellow!25}\textbf{0.961} \\
& P4 & skied daily. Excell & 0.461 \\
\cmidrule(lr){1-4}

\multirow{5}{*}{\textbf{Ex. 3}}
& \cellcolor{blue!10}\textbf{GT} & \cellcolor{blue!10}\emph{screen washer fluid are} & \cellcolor{blue!10}-- \\
& P1 & screen washer software are & \cellcolor{yellow!25}\textbf{0.922} \\
& P2 & screen wipers are changed & 0.641 \\
& P3 & screen washing and dipping & 0.555 \\
& P4 & screen washer fluids & 0.824 \\
\cmidrule(lr){1-4}

\multirow{5}{*}{\textbf{Ex. 4}}
& \cellcolor{blue!10}\textbf{GT} & \cellcolor{blue!10}\emph{bound on the predictive} & \cellcolor{blue!10}-- \\
& P1 & bound on the trace of & \cellcolor{yellow!25}\textbf{0.840} \\
& P2 & -ranked trace- & 0.535 \\
& P3 & bound on  & 0.594 \\
& P4 & bound on P( & 0.582 \\
\cmidrule(lr){1-4}

\multirow{5}{*}{\textbf{Ex. 5}}
& \cellcolor{blue!10}\textbf{GT} & \cellcolor{blue!10}\emph{continues to grow; and} & \cellcolor{blue!10}-- \\
& P1 & reflects wave after wave of & 0.484 \\
& P2 & stretches into space. & \cellcolor{yellow!25}\textbf{0.531} \\
& P3 & heats ocean waters a & 0.461 \\
& P4 & cushions us all from & 0.500 \\
\cmidrule(lr){1-4}

\multirow{5}{*}{\textbf{Ex. 6}}
& \cellcolor{blue!10}\textbf{GT} & \cellcolor{blue!10}\emph{is also convergent according} & \cellcolor{blue!10}-- \\
& P1 & is a convergent integral & \cellcolor{yellow!25}\textbf{0.758} \\
& P2 & shall not be less than & 0.512 \\
& P3 & must have a value, & 0.574 \\
& P4 & is convergent. Because & 0.578 \\

\bottomrule
\end{tabular}
\label{ex:rewards}
\end{table}

\end{document}